\documentclass{article}

\usepackage[nonatbib,preprint]{neurips_2021}

\usepackage[utf8]{inputenc} 
\usepackage[T1]{fontenc}    
\usepackage{hyperref}       
\usepackage{url}            
\usepackage{booktabs}       
\usepackage{amsfonts}       
\usepackage{nicefrac}       
\usepackage{microtype}      
\usepackage{xcolor}         
\usepackage{tablefootnote}
\usepackage[numbers]{natbib}
\usepackage{graphicx}
\usepackage{amssymb,amsmath,amsthm,array}
\usepackage{color}
\usepackage{underoverlap}
\usepackage{subfigure}
\usepackage{multirow}
\usepackage{wrapfig}
\usepackage[rightcaption]{sidecap}
\usepackage{yfonts}
\usepackage{pifont} 
\usepackage{xspace}
\usepackage{tabularx}
\usepackage{tikz}
\usepackage{scalerel}

\usetikzlibrary{arrows.meta,positioning,calc}
\sidecaptionvpos{figure}{c}

\def\b{{\mathbf b}}

\def\e{{\mathbf e}}

\def\i{{\mathbf i}}

\def\p{{\mathbf p}}

\def\s{{\mathbf s}}

\def\u{{\mathbf u}}

\def\x{{\mathbf x}}
\def\y{{\mathbf y}}

\def\A{{\mathbf A}}

\def\D{{\mathbf D}}

\def\HH{{\mathbf H}} 

\def\L{{\mathbf L}}

\def\S{{\mathbf S}}

\def\U{{\mathbf U}}
\def\V{{\mathbf V}}
\def\W{{\mathbf W}}
\def\X{{\mathbf X}}

\def\cN{\mathcal{N}}  
\def\cG{\mathcal{G}}  
\def\cX{\mathcal{X}}  
\def\fX{\mathfrak{X}} 
\def\cE{\mathcal{E}}  
\def\fE{\mathfrak{E}} 
\def\cL{\mathcal{L}} 

\def\cS{\mathcal{S}}  

\newcommand{\Sel}{{\rm \textsc{Sel}}}
\newcommand{\Red}{{\rm \textsc{Red}}}
\newcommand{\Con}{{\rm \textsc{Con}}}
\newcommand{\Pool}{{\rm \textsc{Pool}}}

\newcommand{\UpScale}{{\rm \textsc{UpScale}}}

\newcommand{\argmax}{\operatornamewithlimits{arg\ max}}

\newcommand{\norm}[1]{\left\lVert#1\right\rVert}

\newcommand{\ie}{\emph{i.e.}\xspace}
\newcommand{\eg}{\emph{e.g.}\xspace}
\newcommand{\cf}{\emph{cf.}\xspace}

\newcommand{\etal}{\emph{et al.}\xspace}

\renewcommand{\paragraph}[1]{\textbf{#1\;}}

\definecolor{ForestGreen}{rgb}{0.0, 0.57, 0.13}
\newcommand{\cmark}{\textcolor{ForestGreen}{\ding{51}}}

\newcommand{\codeurl}{\url{https://github.com/danielegrattarola/SRC}}

\title{Understanding Pooling in Graph Neural Networks}

\author{%
  Daniele Grattarola \\
  Università della Svizzera italiana \\
  {\tt\small grattd@usi.ch}\\
  \And
  Daniele Zambon \\
  Università della Svizzera italiana \\
  \And
  Filippo Maria Bianchi \\
  UiT the Arctic University of Norway \\
  NORCE, The Norwegian Research Centre \\
  \And
  Cesare Alippi \\
  Università della Svizzera italiana \\
  Politecnico di Milano \\
}

\begin{document}
\maketitle

\begin{abstract}

Inspired by the conventional pooling layers in convolutional neural networks, many recent works in the field of graph machine learning have introduced pooling operators to reduce the size of graphs. 
The great variety in the literature stems from the many possible strategies for coarsening a graph, which may depend on different assumptions on the graph structure or the specific downstream task.
In this paper we propose a formal characterization of graph pooling based on three main operations, called \emph{selection, reduction,} and \emph{connection}, with the goal of unifying the literature under a common framework. 
Following this formalization, we introduce a taxonomy of pooling operators and categorize more than thirty pooling methods proposed in recent literature. 
We propose criteria to evaluate the performance of a pooling operator and use them to investigate and contrast the behavior of different classes of the taxonomy on a variety of tasks. 
\end{abstract}

\section{Introduction}

Similarly to the convolutional and pooling layers in convolutional neural networks (CNNs), graph neural networks (GNNs) are often built by alternating layers that learn a transformation of the node features and pooling layers that reduce the number of nodes. 
Graph pooling can also be used as an independent operation to produce coarsened representations of given graphs. 

While the techniques for learning node representations have been largely studied, and works like those of \citet{gilmer2017neural} and \citet{battaglia2018relational} have introduced general frameworks to unify the existing literature, less attention has been devoted to pooling layers. Only a few recent works have attempted to systematically analyze the effect of pooling in GNNs~\cite{DBLP:journals/corr/abs-1905-02850,mesquita2020rethinking} and, 
notably, a unifying formulation of pooling operators is still missing. 
In this paper we advance the understanding of graph pooling operators by proposing a universal and modular formalism to study pooling in GNNs and we show what types of operators can be effective in different settings. Specifically:

$\bullet\quad$ We show that graph pooling operators can be seen as the combination of three functions: \emph{selection}, \emph{reduction}, and \emph{connection} (SRC). The selection function groups the nodes of the input graph into subsets called \emph{supernodes}; 
then, the reduction function aggregates each supernode to form an output node and its attributes;
finally, the connection function links the reduced nodes with (possibly attributed) edges and outputs the pooled graph. 
The process is summarized in Figure~\ref{fig:src}.
We also show that the three SRC functions can be interpreted as node- and graph-embedding operations, and that recent theoretical results on the universality of GNNs~\cite{keriven2019universal,maron2019universality} can be exploited to define universal approximators for any pooling operator with continuous SRC functions.

$\bullet\quad$ We propose a comprehensive taxonomy of pooling operators based on specific properties of the SRC functions. In particular, we identify four main properties that characterize pooling operators: a) whether or not the SRC functions are learned, b) whether their complexity is linear or quadratic in the number of nodes, c) whether they produce graphs with a fixed or variable number of nodes, d) whether they pool the graphs hierarchically or globally.

$\bullet\quad$ We identify three evaluation criteria that allow quantifying how much a pooling operator preserves information related to the node attributes, how much it preserves the original topological structure, and how well it can adapt the coarsened graph in relation to the downstream tasks.
Through experiments designed to test how well these criteria are met, we evaluate the performance and analyze the behavior of different classes of pooling operators. 
We provide guidelines to identify the appropriate pooling methods for a given task based on the taxonomy and the fulfillment of the discussed criteria.

$\bullet\quad$ Finally, we release a standardized and framework-agnostic Python API for implementing graph pooling operators.\footnote{\codeurl}
Such general interfaces are the basis on which popular software libraries for creating GNNs~\cite{fey2019fast,grattarola2020graph} are built, as they offer an easy and flexible way to make novel operators available to the research community and to reproduce results.

\def\ggraph{\scalerel{\includegraphics{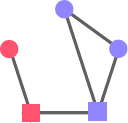}}{\rule[-4\LMpt]{0pt}{32\LMpt}}}
\def\gpool{\scalerel*{\includegraphics{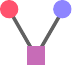}}{\rule[-4\LMpt]{0pt}{20\LMpt}}}
\def\gselone{\scalerel*{\includegraphics{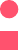}}{\rule[-6\LMpt]{0pt}{16\LMpt}}}
\def\gseltwo{\scalerel*{\includegraphics{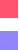}}{\rule[-6\LMpt]{0pt}{16\LMpt}}}
\def\gselthr{\scalerel*{\includegraphics{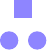}}{\rule[-6\LMpt]{0pt}{16\LMpt}}}
\def\gredone{\scalerel*{\includegraphics{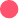}}{G}}
\def\gredtwo{\scalerel*{\includegraphics{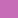}}{G}}
\def\gredthr{\scalerel*{\includegraphics{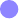}}{G}}
\def\gconone{\scalerel*{\includegraphics{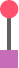}}{\rule[-6\LMpt]{0pt}{16\LMpt}}}
\def\gcontwo{\scalerel*{\includegraphics{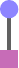}}{\rule[-6\LMpt]{0pt}{16\LMpt}}}
\begin{figure*}[t]
\begin{center}
\tikzstyle{op} = [rectangle, rounded corners,text centered, inner sep=2mm]
\resizebox{\columnwidth}{!}{%
\begin{tikzpicture}[node distance=5mm]
    \node (G) at (0,0) {$\ggraph$};
    \node (sel) [op, draw=blue!50, right=of G] {$\Sel$};
    \node (S) [right=of sel] {$\cS = \left\{ \gselone, \; \gseltwo, \; \gselthr \right\}$};
    \node (red) [op, draw=red!50, above right=0mm and 5mm of S] {$\Red$};
    \node (con) [op, draw=ForestGreen!70, below right=0mm and 5mm of S] {$\Con$};
    \node (X') [right=of red] {$\cX' = \left\{ \gredone, \gredtwo, \gredthr \right\}$};
    \node (E') [right=of con] {$\cE' = \left\{ \gconone, \; \gcontwo \right\}$};
    \node (G') [below right=0mm and 5mm of X'] {$\gpool$};
    
    \tikzstyle{arrow} = [thin,->,>=stealth]
    \draw [arrow] (G) -- (sel);
    \path[draw,>=stealth]
        (sel) edge[->] (S)
        (S)   edge[->,bend left]  (red)
        (S)   edge[->,bend right] (con)
        (red) edge[->]             (X')
        (con) edge[->]             (E')
        (G)   edge[->,bend left]  (red)
        (G)   edge[->,bend right] (con)
        (X'.east)  edge[->,bend left]  (G')
        (E'.east)  edge[->,bend right] (G')
    ;
    
    \node (redbox) [op, draw=red!50, above right=0mm and 5mm of G'] {
        $\x'_k \leftarrow \Red(\cS_k, \cG)$
    };
    \node (conbox) [op, draw=ForestGreen!70, below right=0mm and 5mm of G'] {
        $\e'_{k \rightarrow l}\leftarrow\Con(\cS_k, \cS_l, \cG)$
    };
    
    \draw [thin,dotted] (con.south) --++ (0mm,-3mm) -| (conbox.south);
    \draw [thin,dotted] (red.north) --++ (0mm,3mm) -| (redbox.north);
\end{tikzpicture}
}
\caption{Schematic view of pooling operators: the selection function $\Sel$ groups the nodes into supernodes $\cS_1, \dots, \cS_K$; the reduction function $\Red$ maps each supernode $\cS_k$ to the attribute of node $k$ in the pooled graph; finally, the connection function $\Con$ computes the edges between each pair of new nodes.}
\label{fig:src}
\end{center}
\end{figure*}
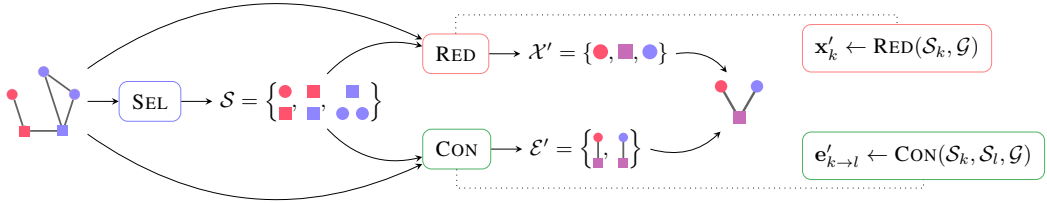

\section{Pooling in graph neural networks}
\label{sec:background}
Among the early uses of graph pooling found in the GNN literature, \citet{bruna2013spectral} mentioned popular graph clustering algorithms~\cite{coates2011selecting,kushnir2006fast,von2007tutorial,dhillon2007weighted} to perform pooling in GNN architectures. In particular, the Graclus algorithm~\cite{dhillon2007weighted} was used by \citet{defferrard2016convolutional} and later adopted in other works on GNNs~\cite{monti2017geometric, levie2017cayleynets,fey2018splinecnn}. 
The literature about learning on point clouds also introduced pooling techniques to generalize the typical pooling layers for grids, with most approaches based on voxelization techniques~\cite{simonovsky2017dynamic,riegler2017octnet,qi2017pointnet++,lei2019octree}.
Many pooling operators have also been proposed based on graph spectral theory~\cite{bianchi2019hierarchical,ma2019graph,hermsdorff2019unifying}, or different clustering, sparsification, and decomposition techniques~\cite{luzhnica2019clique,bacciu2019non,noutahi2019towards,xie2020graph}.

The current trend (and state of the art) in graph pooling has seen the advent of learnable operators that, much like message-passing layers, can dynamically adapt to a particular task to compute optimal pooling. These include clustering approaches like the DiffPool~\cite{ying2018hierarchical} and MinCut~\cite{bianchi2019mincut} operators, as well as a large variety of methods that learn to keep some nodes and discard the others~\cite{graphunet,cangea2018towards,DBLP:journals/corr/abs-1905-02850,lee2019self,ranjan2019asap}. 
Other notable works include those of \citet{diehl2019edge}, \citet{bodnar2020deep}, and a broad group of \emph{global} pooling methods (\cf Section~\ref{sec:taxonomy}) that reduce a whole graph to a single vector~\cite{li2015gated,zhang2018end,wu2019net,navarin2019universal,corcoran2019function,atwood2016diffusion,xu2018powerful,bai2019unsupervised}.

This paper studies such a heterogeneous family of operators to highlight their common characteristics and key differences. 
We start by introducing, in the next section, a definition for graph pooling that encompasses all methods listed above.

\paragraph{Notation} 
We denote an attributed graph with $N$ nodes as a tuple $\cG = (\cX, \cE)$ where $\cX = \{ (i, \x_i) \}_{i = 1:N}$ is the node set, with $\x_i\in\fX$ the attribute of the $i$-th node, and $\cE = \left\{ \big((i,j), \e_{ij}\big)\right\}_{i,j\in 1:N}$ is the edge set, with $\e_{ij}\in\fE$ the attribute associated with the edge between nodes $i$ and $j$. 
With a little abuse of notation, in the following we identify node $(i, \x_i)$ only with its attribute $\x_i$; similarly for edges.
Usually, domains $\fX$ and $\fE$ of the node and edge attributes are real vector spaces, namely, $\fX = \mathbb{R}^F$ and $\fE = \mathbb{R}^H$ for $F, H \in \mathbb{N}$.
Without loss of generality, non-attributed graphs can still fit in this formalism by considering surrogate attributes, such as a constant value or some node-specific property (\eg, the degree).
It is also practical to represent the graph with an adjacency matrix $\A \in \{0, 1\}^{N \times N}$ and a node attribute matrix $\X \in \mathbb{R}^{N \times F}$. In this paper, we consider undirected graphs since the literature mostly focuses on them.

\section{Select, reduce, connect}
\label{sec:src}

Let a graph pooling operator be loosely defined as any function $\Pool$ that maps a graph $\cG$ to a new pooled graph $\cG'=(\cX', \cE')$, with the generic goal of reducing the number of nodes from $N$ to $K < N$. 
To facilitate our study of graph pooling methods, it is useful to isolate the main operations that all methods must perform, regardless of their specific implementation. We identify three such operations: selection, reduction and connection (SRC); see Figure~\ref{fig:src}.
With selection, the operator computes $K$ subsets of nodes, each associated with one node of the output $\cG'$; we refer to them as \emph{supernodes}. With reduction, the operator aggregates the node attributes in each supernode to obtain the node attributes of $\cG'$. Finally, the connection step computes edges among the $K$ reduced nodes. 

The SRC operations allow us to easily describe pooling methods, as done in Table~\ref{tab:implementation}.
Accordingly, we define a pooling operator as any function 
$
    \Pool : \cG \mapsto \cG' = (\cX', \cE')
$
written as the composition of
\begin{gather}
    \underbrace{\cS = \left\{\cS_k\right\}_{k=1:K} = \Sel(\cG);}_{\text{Selection}} \;\;
    \underbrace{\cX'=\left\{\Red( \cG, \cS_k )\right\}_{k=1:K};}_{\text{Reduction}} \;\;
    \underbrace{\cE'=\left\{\Con( \cG, \cS_k, \cS_l )\right\}_{k,l=1:K}.}_{\text{Connection}}
\end{gather}
Different pooling operators are determined by the specific implementations of \Sel, \Red\ and \Con.

\paragraph{Select} The selection function $\Sel$ maps the nodes of the input graph to the nodes of the pooled one. 
The role of $\Sel$ is crucial as it determines the number of nodes in the output graph and what information from the input will be carried over to the output. 
A selection consists of assigning the $N$ input nodes to $K$ sets $\cS_1, \dots, \cS_K \subseteq \cX$, called \emph{supernodes}: $\Sel: \cG \mapsto \cS = \{ \cS_1, \dots, \cS_K \}.$
Each supernode is a set 
$
    \cS_k = \left\{ (\x_i, \s_i) \mid \x_i \in \cX, \s_i \in \mathbb{R}, \s_i > 0 \right\},
$
whose element $(\x_i, \s_i)$ indicates that node $i$ of the input graph impacts on the $k$-th node of the pooled graph. 
The value $\s_i$ is a membership score of node $i$ with respect to supernode $k$, \ie, how much node $i$ contributes to $\cS_k$.
In general, a node can be assigned to zero, one, or multiple supernodes, with different scores.

\paragraph{Reduce} The reduction function computes the node attributes of graph $\cG'$ by aggregating the node attributes of $\cG$ selected in each supernode $\cS_k$.
A reduction consists of applying a function $\Red$ to each supernode $\cS_k$ to produce the $k$-th node attribute $\x_k'$ of $\cG'$: $\Red : \cG, \cS_k \mapsto \x_k' \in \fX.$

\paragraph{Connect} The connection function determines, for each pair of supernodes $\cS_k,\cS_l$, the presence or absence of an edge between the corresponding nodes $k$ and $l$ in the pooled graph. The function also computes the attributes to be assigned to new edges and reads: $\Con :  \cG, \cS_k, \cS_l \mapsto \e_{kl}' \in \fE.$
We assume that the space of edge attributes contains a null attribute encoding the absence of an edge and that the edge set of a graph only contains non-null edges. 

The reason why both $\Red$ and $\Con$ are defined as functions of graph $\cG$ is that their output can depend on the full topology of the input graph in non-trivial ways.
For example, pooling methods based on the graph spectrum often connect the nodes of $\cG'$ based on the whole structure of $\cG$.
However, we notice that many pooling operators implement $\Red$ and $\Con$ as functions of the supernodes only. 

\begin{table*}
    \centering
    \caption{Pooling methods in the SRC framework. $\texttt{GNN}$ indicates a stack of one or more message-passing layers, $\texttt{MLP}$ is a multi-layer perceptron, $\L$ is the normalized graph Laplacian, $\beta$ is a regularization vector (see~\cite{noutahi2019towards}), $\D$ is the degree matrix, $\u_{max}$ is the eigenvector of the Laplacian associated with the largest eigenvalue, $\i$ is a vector of indices, $\A_{\i, \i}$ selects the rows and columns of $\A$ according to $\i$.}
    
    \renewcommand{\arraystretch}{1.5} 
    \resizebox{\textwidth}{!}{%
    \begin{tabular}{@{}llll@{}}
    \toprule
        \textbf{Method}                      & \textbf{Select} & \textbf{Reduce} & \textbf{Connect} \\ \midrule
        DiffPool \cite{ying2018hierarchical} & $\S = \texttt{GNN}_1(\A, \X)$ (w/ auxiliary loss) & $\X' = \S^\top \cdot \texttt{GNN}_2(\A, \X)$ & $\A' = \S^\top \A \S$ \\
        MinCut \cite{bianchi2019mincut}      & $\S = \texttt{MLP}(\X)$ (w/ auxiliary loss) & $\X' = \S^\top \X$ & $\A' = \S^\top \A \S$ \\
        NMF \cite{bacciu2019non}             & Factorize: $\A = \W\HH \, \rightarrow \, \S = \HH^\top$ & $\X' = \S^\top \X$ & $\A' = \S^\top \A \S$ \\
        LaPool \cite{noutahi2019towards}     & $\begin{cases}
                        \V = \norm{\L\X}_d; \\ 
                        \i = \{ i \mid \forall j \in \cN(i): \V_i > \V_j \} \\ 
                        \S = \text{SparseMax}\left( \beta \frac{\X\X_{\i}^\top}{\norm{\X}\norm{\X_{\i}}} \right)%
                   \end{cases}$ & $\X' = \S^\top \X$ & $\A' = \S^\top \A \S$ \\
        Graclus \cite{dhillon2007weighted}   & $\cS_k = \Big\{\x_i, \x_j \mid \argmax_{j} \big( \frac{\A_{ij}}{\D_{ii}} + \frac{\A_{ij}}{\D_{jj}} \big)\Big\}$ & $\X' = \S^\top \X$ & METIS \cite{karypis1997metis} \\
        NDP \cite{bianchi2019hierarchical}   & $\i = \{ i \mid \u_{max, i} > 0 \}$ & $\X' = \X_{\i}$ & Kron r.\ \cite{kron_red}\\
        Top-$K$ \cite{graphunet}             & $\y = \frac{\X\p}{\|\p\|};\,\i = \mathrm{top}_K(\y)$ & $\X' = (\X \odot \sigma(\y))_{\i};$ & $\A' = \A_{\i, \i}$ \\ 
        SAGPool \cite{lee2019self}           & $\y = \texttt{GNN}(\A, \X);\,\i = \mathrm{top}_K(\y)$ & $\X' = (\X \odot \sigma(\y))_{\i};$ & $\A' = \A_{\i, \i}$ \\ 
    \bottomrule
    \end{tabular}
    }
    \label{tab:implementation}
\end{table*}

Given our definition of pooling operators as a combination of the SRC functions, we show in Table~\ref{tab:implementation} how to express several pooling methods proposed in recent literature under the SRC formalism.
We observe that the selection is commonly computed as a matrix $\S \in \mathbb{R}^{N \times K}$, where $\S_{ik}$ indicates the membership score of node $i$ to supernode $k$, and $\S_{ik} = 0$ means that node $i$ is not assigned to supernode $k$.

\paragraph{SRC as embedding operations}
\label{sec:embedding}
Since the three SRC functions are essentially node- and graph-embedding operations, we can rely on well-established theory~\cite{sato2020survey} to study the expressive power of pooling operators when formulated under the SRC framework.

Consider a graph space defined on compact node and edge attribute sets $\fX, \fE$, and let $K(\cG)$ represent the number of nodes of $\cG'=\Pool(\cG)$, where $K(\cG) \le \overline K$ for all $\cG$ and for some finite $\overline K \in \mathbb{N}$. 
By representing the output of the selection function as a matrix $\S \in \mathbb{R}^{N \times \overline K}$, we can then interpret $\Sel$ as permutation-equivariant node embedding operation $\x_i \mapsto \S_{i,:}$, from the space of node attributes to the space of supernodes assignments $\mathbb{R}^{\overline K}$ where we assumed, without loss of generality, that $\S_{i,k}=0$ for all $k > K(\cG)$ (this is necessary to ensure that any number of nodes $K(\cG)$ can be computed by $\Pool$).
Now, let $\cG_{\cS_k}$ indicate an augmentation of $\cG$ such that its node features are $\X_{\cS_k} = \X\|\S_{:,k}$, where $\|$ indicates concatenation, and similarly $\cG_{(\cS_k,\cS_l)}$ is defined by $\X_{(\cS_k,\cS_l)} = \X\|\S_{:,k}\|\S_{:,l}$. It is immediate to see that the augmented graphs are an equivalent way of representing the inputs of $\Red$ and $\Con$, which can then be seen as graph-embedding operations of the form:
\begin{equation}
    \Red: \cG_{\cS_k} \mapsto \x_k'; \;\; \Con: \cG_{(\cS_k,\cS_l)}\mapsto \e'_{kl}.    
\end{equation}
An interesting consequence of this interpretation of SRC is that by implementing $\Sel$ as a universal equivariant network~\cite{keriven2019universal}, and $\Red$ and $\Con$ as universal invariant ones~\cite{maron2019universality}, the resulting operator is a universal approximator for any arbitrary choice of pooling with continuous $\Sel$, $\Red$ and $\Con$. 

\section{Taxonomy of graph pooling}
\label{sec:taxonomy}
The SRC framework is a general template to describe pooling operators and it allows us to characterize the different families of pooling methods that are found in the literature. 
We propose the following taxonomy of pooling operators based on four distinguishing characteristics and we show in Table~\ref{tab:taxonomy} how existing pooling methods fit into this taxonomy. 

\paragraph{Trainability}
\label{sec:trainability}
A first distinction among pooling operators is whether $\Sel$, $\Red$, and $\Con$ are learned end-to-end as part of the overall GNN architecture.
In this case, we say that a method is \emph{trainable}, \ie, the operator has parameters which are learned by optimizing a task-driven loss function, while in all other cases we say that methods are \emph{non-trainable}. 
This distinction is important because, while non-trainable methods are often used as stand-alone algorithms for graph coarsening, trainable methods were specifically designed for GNNs and are a novel research topic of their own.

Generally, non-trainable methods are useful when there is strong prior information about the desired behavior of pooling (\eg, preserving connectivity~\cite{dhillon2007weighted} or filtering out some particular graph frequencies~\cite{noutahi2019towards}). 
These prior assumptions are usually grounded on graph-theoretical properties and are useful when few data are available, since they do not increase the overall number of parameters and introduce no additional optimization objectives when training the GNN. 
A well-known example of non-trainable pooling is the conventional grid pooling of CNNs, which pools spatially localized groups of pixels.
On the other hand, trainable methods are more flexible and make fewer assumptions about the desired result. 
Therefore, they are useful in problems where the best pooling strategy is not known \emph{a priori}. 
However, note that it is possible to integrate priors about the desired pooling behavior also in trainable methods (\eg, the MinCut~\cite{bianchi2019mincut} operator also optimizes a normalized cut objective to ensure that supernodes are homogeneous).
These additional assumptions usually act as a regularization.

\begin{figure*}
\centering
\hfill%
\begin{minipage}{.65\textwidth}
    \centering
    \subfigure[]{\includegraphics[width=.45\linewidth]{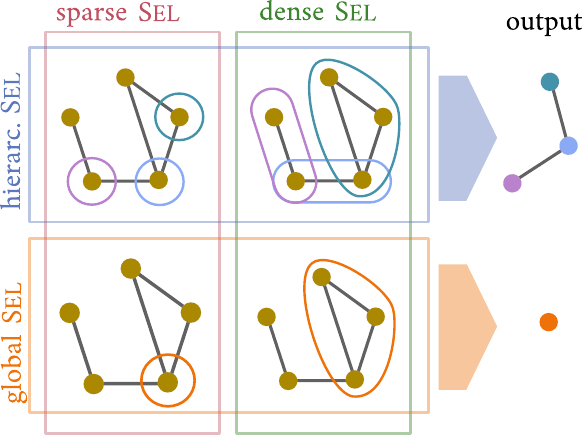}\label{fig:pool-a}}\hspace{0.5cm}
    \subfigure[]{\includegraphics[width=.37\linewidth]{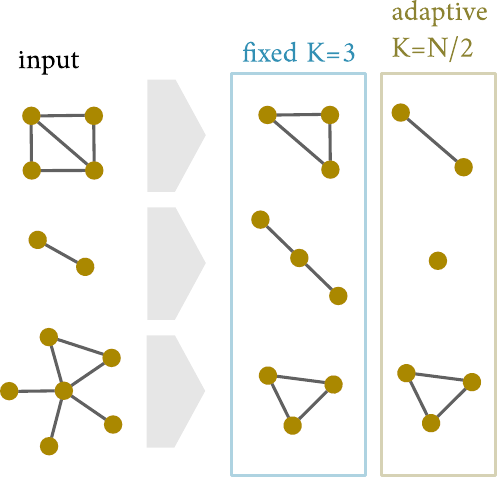}\label{fig:pool-b}}
    \caption{\protect\ref{fig:pool-a} Sparse supernodes have a constant cardinality ($|\cS_k|=1$) while dense supernodes scale with the size of the graph. Hierarchical methods reduce the graph gradually, while global methods always return one node. \protect\ref{fig:pool-b} Fixed methods return the same number of nodes ($K=3$) while adaptive methods return graphs of size proportional to the input.
    }
    \label{fig:desiderata}
\end{minipage}~\hfill~%
\begin{minipage}{.3\textwidth}
    \vspace{.8cm}
    \centering
    \includegraphics[width=.9\linewidth]{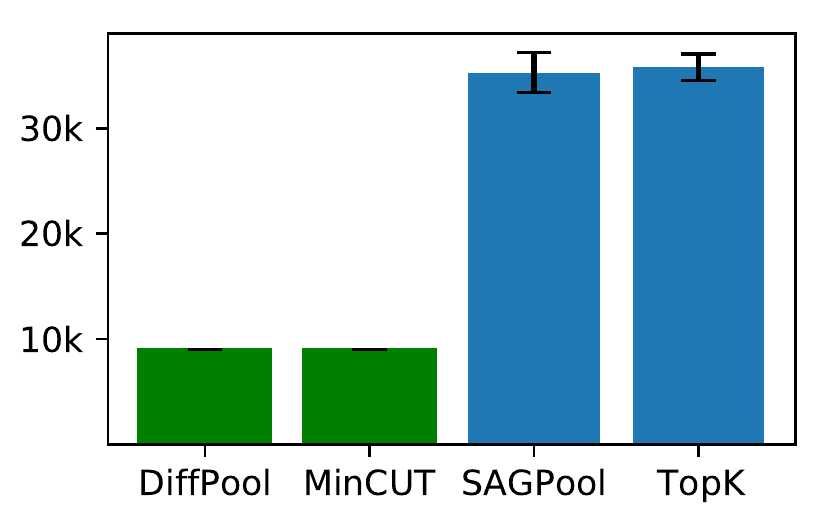}
    \vspace{.4cm}
    \caption{Maximum number of nodes that can be processed by dense (in green) and sparse (in blue) methods.}
    \label{fig:memory}
\end{minipage}\hfill~
\end{figure*}

\paragraph{Density of the supernodes}
\label{sec:density}
A second axis of the taxonomy is concerned with the size of supernodes and the consequent cost of computing the selection function. We define the \emph{density} of a pooling operator as the expected value $\mathbb{E}\left[|\cS_k|/N\right]$ of the ratio between the cardinality of a supernode $\cS_k$ and the number of nodes in $\cG$. 
We say that a method is \emph{dense} if $\Sel$ generates supernodes $\cS_k$ whose cardinality is $O(N)$, and \emph{sparse} if supernodes have constant cardinality $O(1)$.\footnote{Intermediate situations---\eg, $O(\log N)$---are also possible, although here we focus on the two limit cases of $O(N)$ and $O(1)$.} Fig.~\ref{fig:pool-a} shows an example of sparse and dense selection.

This distinction is key, since sparse methods require much less computational resources, especially in terms of memory, which is a significant bottleneck even in modern GPUs. This makes them scale better to large graphs. However, as we show in Section~\ref{sec:eval}, sparse selection is a harder operation to learn than dense selection, and may result in unexpected behaviors.

\paragraph{Adaptability of $K$}
\label{sec:adaptability}
It is also possible to distinguish pooling methods according to the number of nodes $K$ of the pooled graph.
If $K$ is constant and independent from the input graph size, we say that a pooling method is \emph{fixed}. In this case, $K$ is a hyperparameter of the pooling operator and the output graph will always have $K$ nodes. 
For example, $K$ can be the number of output features of a neural network used to compute cluster assignments~\cite{bianchi2019mincut,ying2018hierarchical}.
On the other hand, if the number of supernodes is a function $K(\cG)$ of the input graph we say that the method is \emph{adaptive}. 
In many cases, $K(\cG)$ is a function of $N$ (\eg the ratio $N/2$), but $K(\cG)$ could also depend on the input graph in a more complex way (\eg,~\cite{noutahi2019towards,DBLP:journals/corr/abs-1905-02850}).
 
Adaptive pooling methods can compute graphs that have a size proportional to that of the input. On the other hand, all the coarsened graphs generated by fixed methods will have the same size.
This can lead to situations where $K > N$ for some graphs, causing them to be upscaled by pooling, rather than coarsened.
Fig.~\ref{fig:pool-b} compares fixed and adaptive pooling and shows an example (2\textsuperscript{nd} row) where fixed pooling upscales the graph.
For data with a wide or skewed distribution of the number of nodes, the values commonly chosen for $K$ in fixed methods, like the average number of nodes in the training set, may cause several small graphs to be upscaled and very big graphs to be excessively shrunk. 
Therefore, if the relative graph size is important for solving a particular task, adaptive methods should be preferred.

\paragraph{Hierarchy}
\label{sec:hierarchy}
A distinction often found in the literature is that between ``regular'' and global pooling, which is extremely evident, to the point where global pooling is usually referred to as a separate operation called ``readout''. Here we show that such a distinction can be formalized with the SRC framework.
Specifically, \emph{global pooling} indicates those methods that reduce a graph to a single node, discarding all topological information.
A pooling method is global if it is fixed with $K=1$, \ie, it returns a degenerate single-node graph represented by its attribute. Also, the connection function is a constant map to the empty set. 
On the other hand, we indicate all other methods as \emph{hierarchical} pooling operators. Fig.~\ref{fig:pool-a} shows an example of hierarchical and global pooling.

Hierarchical and global pooling operators have different roles and both can be part of the same GNN architecture for graph-level learning. The former provide a multi-resolution representation of the graph from which the GNN can gradually distill high-level properties, while the latter compute graph embeddings to interface with traditional layers operating on vectors.

{
\newcolumntype{Y}{>{\arraybackslash}X}
\newcommand{\colh}[1]{\textbf{#1}}
\renewcommand{\arraystretch}{1}
\begin{table*}
    \caption{Taxonomy of pooling operators. Methods are divided in trainable or non-trainable (\textbf{T / nT}), dense or sparse (\textbf{D / S}), fixed or adaptive (\textbf{F / A}), and hierarchical or global (\textbf{H / G}).}
    \centering
    \resizebox{\textwidth}{!}{%
    \begin{tabular}{@{}l|cc|cc|cc|cc@{}}
        \toprule
        \colh{Method} & \colh{T} & \colh{nT} & \colh{D} & \colh{S} & \colh{F} & \colh{A} & \colh{H} & \colh{G} \\ \midrule
        DiffPool \cite{ying2018hierarchical}, MinCut \cite{bianchi2019mincut}, StructPool \cite{yuan2020structpool} & \cmark &  & \cmark &  & \cmark &  & \cmark &  \\
        Top-$K$ methods \cite{graphunet,cangea2018towards,lee2019self,DBLP:journals/corr/abs-1905-02850,ranjan2019asap,zhang2021hierarchical}, Edge Contract.\ \cite{diehl2019edge} & \cmark &  &  & \cmark &  & \cmark & \cmark &  \\ \midrule
        Coates and Ng \cite{coates2011selecting}, Voxelization-based \cite{simonovsky2017dynamic,riegler2017octnet,qi2017pointnet++,lei2019octree}  &  & \cmark & \cmark &  &  & \cmark & \cmark &  \\
        NMF \cite{bacciu2019non}, EigenPooling \cite{ma2019graph}, LaPool \cite{noutahi2019towards}, Clique \cite{luzhnica2019clique}  &  & \cmark & \cmark &  &  & \cmark & \cmark &  \\
        Xie \etal\cite{xie2020graph}, MPR \cite{bodnar2020deep}  &  & \cmark & \cmark &  &  & \cmark & \cmark &  \\
        Graclus \cite{dhillon2007weighted}, NDP \cite{bianchi2019hierarchical}, Pooling in CNNs &  & \cmark &  & \cmark &  & \cmark & \cmark &  \\ \midrule
        \cite{li2015gated,vinyals2015order,navarin2019universal} & \cmark &  & \cmark &  & \cmark &  &  & \cmark \\
        \cite{zhang2018end,wu2019net,corcoran2019function,atwood2016diffusion,xu2018powerful,bai2019unsupervised} &  & \cmark & \cmark &  & \cmark &  &  & \cmark \\
        Scarselli \etal\cite{scarselli2009graph} &  & \cmark &  & \cmark & \cmark &  &  & \cmark \\
        \bottomrule
    \end{tabular}%
}
    \label{tab:taxonomy}
\end{table*}%
}

\paragraph{Discussion} The main differences among pooling methods are in the selection function, while much less variety is found in the reduction and connection functions.
A majority of methods (\eg, \cite{simonovsky2017dynamic,qi2017pointnet++,bacciu2019non,luzhnica2019clique,ma2019graph,noutahi2019towards}) have adaptive $K$, with a dense and non-trainable selection.
Adaptive methods are the most commonly found in the literature, although fixed pooling operators are currently the state of the art~\cite{ying2018hierarchical,bianchi2019mincut}. We also note that, to the best of our knowledge, there are no pooling operators that are trainable, dense, and adaptive, which could be an interesting research topic in the near future.

Considering density and adaptability, we see that the memory cost of a pooling operator can range from $O(1)$ (sparse and fixed) to $O(N^2)$ (dense and adaptive).
This is especially relevant for trainable methods, which usually need to fit into memory-bound computational units like GPUs and TPUs. Figure~\ref{fig:memory} shows the maximum number of nodes that can be processed by the trainable methods considered in Section~\ref{sec:eval}, without causing a GPU-out-of-memory exception (details in the appendix). 
As expected, sparse methods can pool graphs up to four times bigger than dense ones.

\section{Evaluation}
\label{sec:eval}
We argue that there is no single general-purpose measure to quantify the performance of a graph pooling algorithm and the quality of a coarsened graph.
In this section, we define three evaluation criteria for pooling operators and design experiments to test whether different classes of methods are able to meet them.
In particular, we evaluate operators based on their ability to 1) preserve the information content of the node attributes, 2) preserve the topological structure, and 3) preserve the information required to solve various classification tasks.
Our goal is to contrast the categories of the proposed taxonomy according to these three criteria.
To perform the comparison, we consider the eight hierarchical pooling methods of Table~\ref{tab:implementation} as representatives:
MinCut~\cite{bianchi2019mincut} and DiffPool~\cite{ying2018hierarchical} are trainable, dense and fixed; Top-$K$~\cite{graphunet,cangea2018towards} and SAGPool~\cite{lee2019self} are trainable, sparse, and adaptive; NMF~\cite{bacciu2019non} and LaPool~\cite{noutahi2019towards} are non-trainable, dense and adaptive; Graclus~\cite{dhillon2007weighted} and NDP~\cite{bianchi2019hierarchical} are non-trainable, sparse and adaptive. 
All implementation details are in the appendix. 

\paragraph{Preserving node attributes}
\label{sec:exp1}
As a first experiment we test the ability of pooling methods to preserve node information.
We consider the task of reconstructing the original coordinates of a geometric point cloud from its pooled version.
We configure a graph autoencoder to pool the node attributes and then lift them back to the original size using an appropriate lift operator for each method (details in the appendix). 
Note that this experiment evaluates the quality of the pooling methods in compressing node information, but it does not test their generalization capability since the autoencoder is independently fit on each point could.

\begin{figure}
    \centering
    \includegraphics[width=\textwidth]{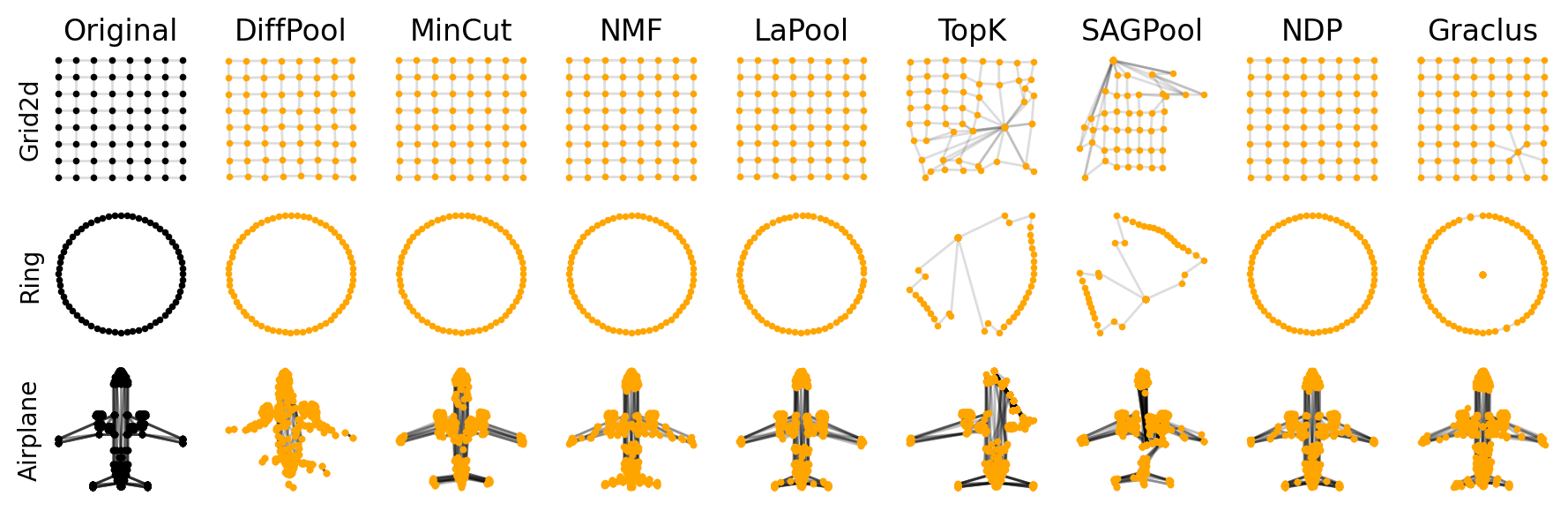}
    \caption{Node attributes (point locations) reconstructed with different operators in the autoencoder experiment.}
    \label{fig:ae_rec_graphs}
\end{figure}

Table~\ref{tab:results_point_cloud} reports the average and standard deviation of the mean squared error (MSE) obtained by the eight methods on different point clouds from the PyGSP library~\cite{pygsp} and the ModelNet40 dataset~\cite{wu20153d}. 
Figure~\ref{fig:ae_rec_graphs} contrasts the original point cloud with the points reconstructed from each pooling method while the connectivity is unchanged; the figures for all point clouds are available in the supplementary material. 
As baseline values for the MSE, we report the mean squared distance between adjacent points: $\gamma = (|\cE|\,F)^{-1} \sum_{(i,j)\in\cE} |\x_i - \x_j|_2^2$; the intuition behind this baseline is that reconstructed points with a MSE score larger than the reference $\gamma$ cannot be matched, on average, with the original points.
We observe that, as the point clouds grow in size, many operators cannot achieve MSE $< \gamma$ ({\color{red!60!black}red entries}). The two methods that stand out from Table~\ref{tab:results_point_cloud} are the non-trainable NMF and NDP, as confirmed also by their respective average ranks across datasets. 

\begin{table}[]
\centering
\caption{MSE (values in scale of $10^{-3}$) in the autoencoder experiment.\protect\footnotemark\  The \textbf{Rank} row indicates the average ranking of the methods across all datasets.}
\label{tab:results_point_cloud}
\resizebox{\textwidth}{!}{%
\newcommand{\outv}[1]{{\color{red!60!black}#1}}
\begin{tabular}{@{}llllllllll@{}}
\toprule
{} & \textit{Ref. $\gamma$} &       \textbf{DiffPool} &          \textbf{MinCut} &             \textbf{NMF} &          \textbf{LaPool} &            \textbf{TopK} &         \textbf{SAGPool} &             \textbf{NDP} &         \textbf{Graclus} \\
\midrule
\textbf{Grid2d   } & \textit{\scriptsize 7.812} &  0.010 \tiny{$\pm$0.005} &  0.002 \tiny{$\pm$0.002} &  \textbf{0.000 \tiny{$\pm$0.000}} &  0.002 \tiny{$\pm$0.001} &  \outv{18.86} \tiny{$\pm$3.923} &  \outv{16.61} \tiny{$\pm$3.270} &  \textbf{0.000 \tiny{$\pm$0.000}} &  0.109 \tiny{$\pm$0.000} \\
\textbf{Ring     } & \textit{\scriptsize 4.815} &  0.018 \tiny{$\pm$0.003} &  0.001 \tiny{$\pm$0.000} &  \textbf{0.000 \tiny{$\pm$0.000}} &  0.052 \tiny{$\pm$0.046} &  \outv{132.2} \tiny{$\pm$4.133} &  \outv{148.5} \tiny{$\pm$30.10} &  \textbf{0.000 \tiny{$\pm$0.000}} &  0.600 \tiny{$\pm$0.000} \\
\textbf{Bunny    } & \textit{\scriptsize 6.874} &  3.901 \tiny{$\pm$0.275} &  \textbf{0.208 \tiny{$\pm$0.034}} &  0.339 \tiny{$\pm$0.055} &  0.610 \tiny{$\pm$0.103} &  \outv{15.32} \tiny{$\pm$3.557} &  \outv{16.10} \tiny{$\pm$1.722} &  0.373 \tiny{$\pm$0.070} &  0.332 \tiny{$\pm$0.043} \\
\textbf{Airplane } & \textit{\scriptsize 0.097} &  0.094 \tiny{$\pm$0.022} &  0.005 \tiny{$\pm$0.002} &  0.020 \tiny{$\pm$0.000} &  \textbf{0.002 \tiny{$\pm$0.000}} &  0.096 \tiny{$\pm$0.028} &  \outv{0.268} \tiny{$\pm$0.081} &  0.012 \tiny{$\pm$0.000} &  0.009 \tiny{$\pm$0.000} \\
\textbf{Car      } & \textit{\scriptsize 0.028} &  \outv{0.143} \tiny{$\pm$0.127} &  \outv{0.535} \tiny{$\pm$0.200} &  0.016 \tiny{$\pm$0.001} &                      \outv{OOR} &  \outv{0.229} \tiny{$\pm$0.023} &  \outv{0.204} \tiny{$\pm$0.029} &  \textbf{0.009 \tiny{$\pm$0.000}} &  \outv{0.102} \tiny{$\pm$0.000} \\
\textbf{Guitar   } & \textit{\scriptsize 0.091} &  \outv{0.101} \tiny{$\pm$0.025} &  \outv{0.313} \tiny{$\pm$0.000} &  0.007 \tiny{$\pm$0.000} &                      \outv{OOR} &  0.056 \tiny{$\pm$0.051} &  0.060 \tiny{$\pm$0.044} &  \textbf{0.005 \tiny{$\pm$0.000}} &  0.010 \tiny{$\pm$0.000} \\
\textbf{Person   } & \textit{\scriptsize 0.013} &  \outv{0.077} \tiny{$\pm$0.041} &  \outv{0.301} \tiny{$\pm$0.000} &  \textbf{0.001 \tiny{$\pm$0.000}} &                      \outv{OOR} &  \outv{0.055} \tiny{$\pm$0.012} &  \outv{0.062} \tiny{$\pm$0.033} & \textbf{0.001 \tiny{$\pm$0.000}} & \textbf{0.001 \tiny{$\pm$0.000}} \\
\midrule
\textbf{Rank    } & &                     5.29 &                     4.29 &                     2.14 &                     5.43 &                     6.14 &                     6.57 &                     \textbf{1.86} &                     3.43 \\
\bottomrule
\end{tabular}
}
\end{table}

\begin{figure}
    \centering
    \includegraphics[width=\textwidth]{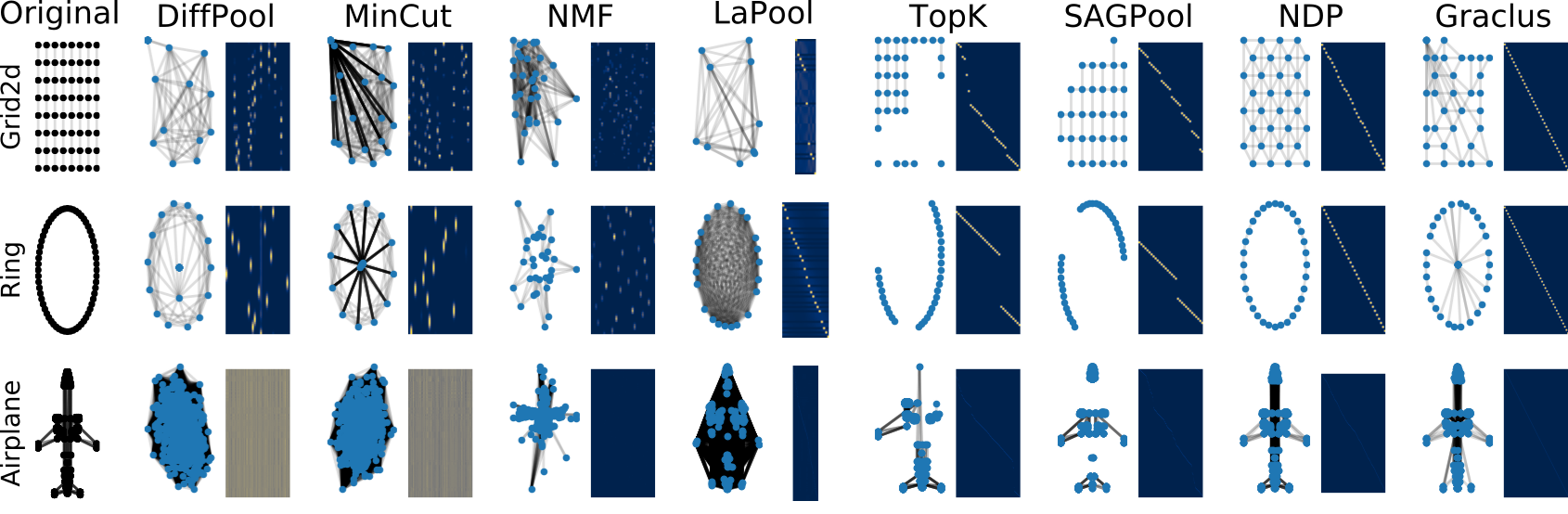}
    \caption{Graphs pooled with different operators in the autoencoder experiment with the modified \Red\ function, and the associated selection matrices $\S$.
    }
    \label{fig:ae_select_graphs}
\end{figure}

\emph{Interpreting the selection operation.}
\label{sec:exp_select}
Figure~\ref{fig:ae_select_graphs} depicts a variant of the coarsened graphs where the \Sel\ and \Con\ operations are the same as in Table~\ref{tab:implementation}, but the \Red\ function is replaced by 
\begin{equation}
\label{eq:modified_red}
    \X' =\S^\top \X.
\end{equation}
This modification is crucial to interpret the \Sel\ operation because most of the pooling methods use message passing layers before the reduction (see the autoencoder architecture details), which makes the node feature space $\fX'$ not directly comparable with the original $2$ or $3$-dimensional input space. Conversely, the reduction in \eqref{eq:modified_red} gives (weighted) averages of the supernodes with the benefits of maintaining points in the input space and locating them in the supernodes' centres of mass. 

Two main patterns emerge. 
First, we see that the two non-trainable sparse methods (NDP and Graclus) perform a rather uniform node subsampling, in such a way that the reduced node features are good representative of the original input, which may facilitate the reconstruction of the input node features, as confirmed by the low MSE in Table~\ref{tab:results_point_cloud}.
Second, trainable and sparse methods (TopK and SAGPool) tend to cut off entire portions of the graphs, therefore discarding essential node information.

\begin{figure}
    \centering
    \includegraphics[width=\textwidth]{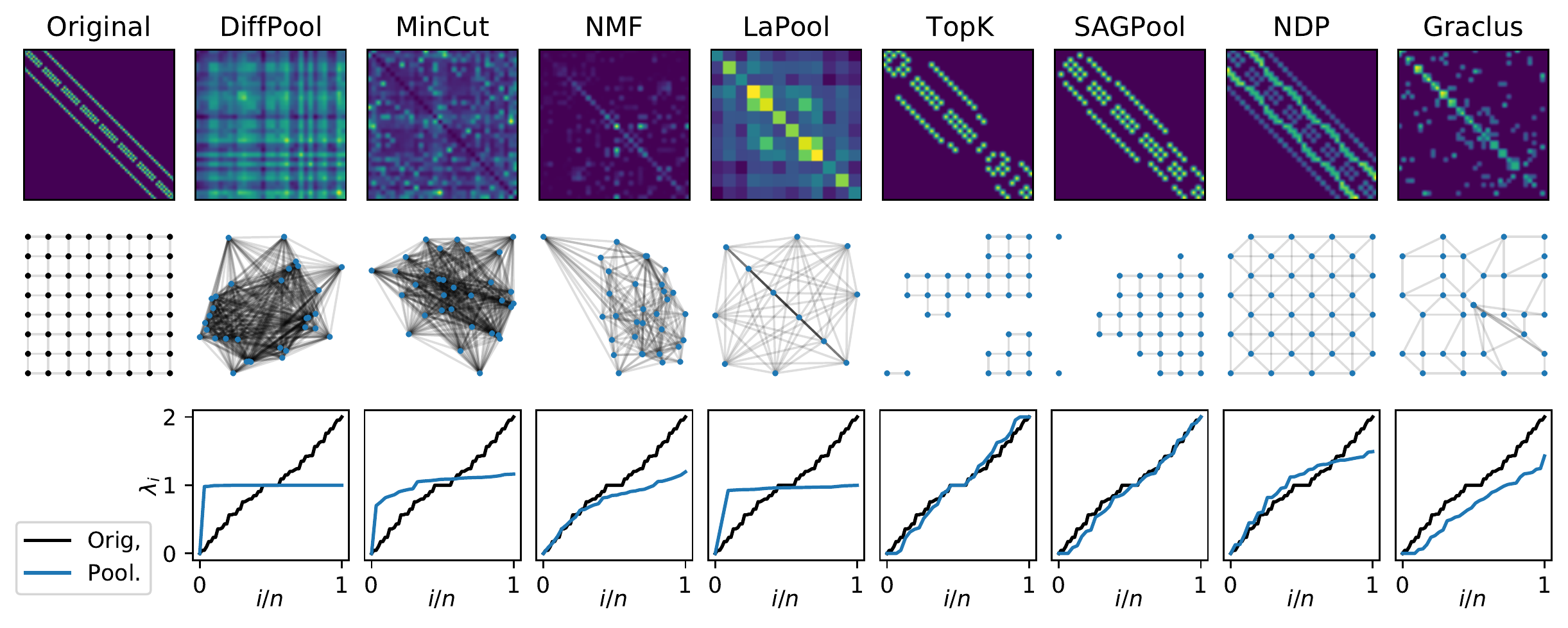}
    \caption{Results on a regular grid when optimizing for spectral similarity. Top: the coarsened adjacency matrices. Middle: the coarsened graphs with modified \Red\ function. Bottom: the eigenvalues of the normalized Laplacian before (black) and after (blue) pooling. The indices of the eigenvalues are rescaled to fill $[0,1]$.}
    \label{fig:spectral_sim_grid}
\end{figure}

\paragraph{Preserving structure}
\label{sec:spectral_sim}
In this experiment we study the structural similarity between the input and coarsened graphs $\cG$ and $\cG'$, respectively, by comparing the quadratic forms associated with their respective combinatorial Laplacian matrices $\L$ and $\L'$. This evaluation criterion has also been recently studied by \citet{loukas2019graph}, \citet{hermsdorff2019unifying}, and \citet{cai2021graph}, and allows us to compare graphs of different sizes. Specifically, we consider the quadratic loss 
$\cL(\cG, \cG') = \sum_{i=0}^d \| \X_{:, i}^{\top} \L \X_{:, i} - {\X'}_{:, i}^{\top} \L' \X'_{:, i} \|,$ 
where $\X$ is an arbitrary graph signal and $\X'$ its reduction. In this experiment, we choose $\X$ to be the concatenation of the first 10 eigenvectors of $\L$ and the node coordinates of $\cG$; all columns are $\ell_2$-normalized.
For trainable methods, we directly minimize the loss as a self-supervised target.
Table~\ref{tab:spectral_sim} reports the average loss obtained by the eight operators on different graphs from the PyGSP library, while Figure~\ref{fig:spectral_sim_grid} shows examples of pooled graphs and their spectra. We show the result for Grid2d since it is easier to interpret visually; the figures for all graphs are available in the supplementary material.

Trainable dense methods can generate coarsened graphs with a quadratic loss w.r.t.\ the original graph lower than their non-trainable or sparse counterparts. 
Interestingly, from the bottom row of Figure~\ref{fig:spectral_sim_grid} we see that a low quadratic loss does not necessarily imply a good alignment of the spectra. 
For example, on the regular grid in Figure~\ref{fig:spectral_sim_grid}, the excellent spectral alignment achieved by Top-K and SAGPool is not reflected by a low quadratic loss value (0.596 and 0.361 respectively).
While in principle this experiment focuses on comparing only \Sel\ and \Con, we are also evaluating \Red\ since it affects the loss that depends on $\X'$. This can explain the discrepancy between the loss values and the eigenvalues plots.

\emph{Properties of the connection operation.}
\label{sec:exp_connect}
From Figure~\ref{fig:spectral_sim_grid}, we see that dense methods (DiffPool, MinCut, NMF, LaPool) yield coarsened graphs that are densely connected. We can make a similar observation for the autoencoder experiment with modified \Red\ operation in Eq.~\eqref{eq:modified_red}, as shown in Figure~\ref{fig:ae_select_graphs}.
However, in these dense graphs most of the edge weights are also small. This is quantitatively reported in Table~\ref{tab:ss_density}, in which we compare the density of edges ($|\cE'| / K^2$) and the median edge weight of the coarsened graphs for Grid2d, Minnesota and Sensor. An extended version of this table is reported in the supplementary material.

\begin{table}[]
\centering
\caption{Average quadratic loss in the spectral similarity experiment.
}
\label{tab:spectral_sim}
\resizebox{\textwidth}{!}{%
\begin{tabular}{@{}lllllllll@{}}
\toprule
{} &        \textbf{DiffPool} &          \textbf{MinCut} &             \textbf{NMF} &          \textbf{LaPool} &            \textbf{TopK} &         \textbf{SAGPool} &             \textbf{NDP} &         \textbf{Graclus} \\
\midrule
\textbf{Grid2d   } &  \textbf{0.002 \tiny{$\pm$0.000}} &  0.099 \tiny{$\pm$0.016} &  0.369 \tiny{$\pm$0.000} &  8.486 \tiny{$\pm$0.000} &  0.483 \tiny{$\pm$0.001} &  0.306 \tiny{$\pm$0.017} &  0.068 \tiny{$\pm$0.000} &  0.375 \tiny{$\pm$0.000} \\
\textbf{Ring     } &  0.001 \tiny{$\pm$0.000} &  \textbf{0.000 \tiny{$\pm$0.000}} &  0.050 \tiny{$\pm$0.000} &  5.603 \tiny{$\pm$0.000} &  0.067 \tiny{$\pm$0.000} &  0.034 \tiny{$\pm$0.001} &  0.002 \tiny{$\pm$0.000} &  0.058 \tiny{$\pm$0.000} \\
\textbf{Sensor   } &  \textbf{0.010 \tiny{$\pm$0.000}} &  0.155 \tiny{$\pm$0.005} &  1.177 \tiny{$\pm$0.000} &  28.99 \tiny{$\pm$0.000} &  1.306 \tiny{$\pm$0.001} &  0.721 \tiny{$\pm$0.077} &  0.486 \tiny{$\pm$0.000} &  1.027 \tiny{$\pm$0.000} \\
\textbf{Bunny    } &  \textbf{0.011 \tiny{$\pm$0.003}} &  0.272 \tiny{$\pm$0.013} &  40.48 \tiny{$\pm$0.000} &                  $>10^3$ &  1.251 \tiny{$\pm$0.000} &  0.708 \tiny{$\pm$0.138} &  0.156 \tiny{$\pm$0.000} &  1.228 \tiny{$\pm$0.000} \\
\textbf{Minnes.} &  \textbf{0.000 \tiny{$\pm$0.000}} &  0.004 \tiny{$\pm$0.000} &  7.117 \tiny{$\pm$0.000} &  4.030 \tiny{$\pm$0.000} &  0.004 \tiny{$\pm$0.000} &  0.001 \tiny{$\pm$0.000} &  \textbf{0.000 \tiny{$\pm$0.000}} &  0.080 \tiny{$\pm$0.000} \\
\textbf{Airfoil  } &  \textbf{0.000 \tiny{$\pm$0.000}} &  0.006 \tiny{$\pm$0.000} &  2.604 \tiny{$\pm$0.000} &  26.97 \tiny{$\pm$0.000} &  0.006 \tiny{$\pm$0.000} &  0.003 \tiny{$\pm$0.000} &  \textbf{0.000 \tiny{$\pm$0.000}} &  0.048 \tiny{$\pm$0.000} \\
\midrule
\textbf{Rank     } &                     \textbf{1.17} &                     2.83 &                     6.33 &                     7.83 &                     5.83 &                     3.67 &                     2.00 &                     5.67 \\

\bottomrule
\end{tabular}
}
\end{table}

\begin{table}[]
\centering
\caption{Density and median weight of the edges of the coarsened graphs in the spectral similarity experiment.}
\label{tab:ss_density}
\scriptsize
\resizebox{\textwidth}{!}{%
\begin{tabular}{@{}lllllllllll@{}}
\toprule
       &        &    \textbf{Original} & \textbf{DiffPool} &      \textbf{MinCut} & \textbf{NMF} & \textbf{LaPool} & \textbf{TopK} & \textbf{SAGPool} & \textbf{NDP} & \textbf{Graclus} \\
\midrule
\multirow{2}{*}{\textbf{Grid2d}} & \textbf{Density} &                0.055 &             0.969 &                0.969 &        0.463 &           0.917 &         0.084 &            0.092 &        0.189 &            0.103 \\
       & \textbf{Median} &                1.000 &             0.216 &                0.024 &        0.018 &           1.445 &         1.000 &            1.000 &        0.500 &            0.154 \\
\midrule
\multirow{2}{*}{\textbf{Minnes.}} & \textbf{Density} &  9.47$\cdot 10^{-4}$ &             0.999 &                0.999 &        0.010 &           0.999 &         0.002 &            0.002 &        0.003 &            0.002 \\
       & \textbf{Median} &                1.000 &             0.004 &  7.58$\cdot 10^{-4}$ &        0.014 &           0.013 &         1.000 &            1.000 &        0.333 &            0.204 \\
\midrule
\multirow{2}{*}{\textbf{Sensor}} & \textbf{Density} &                0.159 &             0.969 &                0.969 &        0.844 &           0.875 &         0.273 &            0.230 &        0.529 &            0.227 \\
       & \textbf{Median} &                0.742 &             0.463 &  2.42$\cdot 10^{-4}$ &        0.005 &           6.147 &         0.765 &            0.756 &        0.201 &            0.103 \\
\bottomrule
\end{tabular}
}
\end{table}

\footnotetext{\label{note}``OOR" indicates Out Of Resources, \ie, either we could not fit a batch size of 8 graphs on an Nvidia Titan V GPU or it took more than 24 hours to complete training. Values of $0.000$ indicate any value $<10^{-6}$.}

\paragraph{Preserving task-specific information}
\label{sec:exp_task}
In our final experiment we consider several benchmarks of graph classification to test the third criterion. A high classification accuracy implies that an operator can selectively preserve information based on the requirements of the task at hand. We consider graph classification problems from the TUDataset~\cite{morris2020tudataset}, the ModelNet10 dataset~\cite{wu20153d}, and the Colors-3 and Triangles datasets introduced by \citet{DBLP:journals/corr/abs-1905-02850}. 

\newcommand{\grh}[1]{{\color{green!50!black}#1}}
\setlength{\intextsep}{2pt}%
\begin{table}
\centering
\caption{Accuracy on the graph classification benchmarks.\textsuperscript{\ref{note}}}
\resizebox{\textwidth}{!}{%
\begin{tabular}{@{}llllllllll@{}}
\toprule
{} &     \textit{No-pool} &     \textbf{DiffPool} &       \textbf{MinCut} &          \textbf{NMF} &       \textbf{LaPool} &          \textbf{TopK} &       \textbf{SAGPool} &          \textbf{NDP} &      \textbf{Graclus} \\
\midrule
\textbf{Colors-3 } & 40.8\tiny{$\pm$2.1} & \grh{55.2} \tiny{$\pm$1.5} &  \textbf{\grh{60.1} \tiny{$\pm$4.0}} &  29.7 \tiny{$\pm$1.7} &  \grh{44.9} \tiny{$\pm$1.0} &   26.9 \tiny{$\pm$4.0} &   34.4 \tiny{$\pm$5.2} &  25.4 \tiny{$\pm$1.8} &  29.5 \tiny{$\pm$2.0} \\
\textbf{Triangles} & 93.5\tiny{$\pm$0.7} &  91.3 \tiny{$\pm$0.2} &  \textbf{\grh{95.3} \tiny{$\pm$0.5}} &  58.1 \tiny{$\pm$5.2} &  88.8 \tiny{$\pm$0.8} &  75.2 \tiny{$\pm$17.3} &   80.3 \tiny{$\pm$8.6} &  75.3 \tiny{$\pm$1.0} &  71.4 \tiny{$\pm$1.7} \\
\textbf{Proteins } & 68.8\tiny{$\pm$2.8} &  \grh{70.0} \tiny{$\pm$0.6} &  \textbf{\grh{73.8} \tiny{$\pm$0.8}} &  68.9 \tiny{$\pm$3.4} &  \grh{72.9} \tiny{$\pm$2.0} &   \grh{71.3} \tiny{$\pm$0.8} &   \grh{73.7} \tiny{$\pm$0.8} &  68.4 \tiny{$\pm$3.4} &  \grh{72.6} \tiny{$\pm$1.1} \\
\textbf{Enzymes  } & 83.6\tiny{$\pm$2.0} &  72.4 \tiny{$\pm$3.9} &  83.6 \tiny{$\pm$0.6} &  32.4 \tiny{$\pm$8.1} &  \grh{85.0} \tiny{$\pm$1.2} &   81.0 \tiny{$\pm$0.4} &  68.8 \tiny{$\pm$18.8} & \grh{84.8} \tiny{$\pm$3.2} &  \textbf{\grh{85.4} \tiny{$\pm$4.1}} \\
\textbf{DD       } & 81.1\tiny{$\pm$0.4} &  75.6 \tiny{$\pm$1.8} &  \textbf{\grh{82.5} \tiny{$\pm$0.9}} &                   OOR &                   OOR &   80.4 \tiny{$\pm$0.9} &   79.0 \tiny{$\pm$2.7} &  79.6 \tiny{$\pm$1.2} &  78.3 \tiny{$\pm$2.9} \\
\textbf{Mutagen. } & \textbf{78.0\tiny{$\pm$1.6}} &  76.2 \tiny{$\pm$1.4} &  73.9 \tiny{$\pm$1.6} &  70.3 \tiny{$\pm$1.6} &  75.3 \tiny{$\pm$0.1} &   75.8 \tiny{$\pm$1.4} &   76.9 \tiny{$\pm$1.4} &  76.9 \tiny{$\pm$1.0} &  74.2 \tiny{$\pm$0.5} \\
\textbf{ModelNet } & 81.0\tiny{$\pm$0.5} &  70.4 \tiny{$\pm$2.4} &  75.9 \tiny{$\pm$1.2} &                   OOR &                   OOR &   74.1 \tiny{$\pm$3.0} &   71.9 \tiny{$\pm$2.6} &  77.1 \tiny{$\pm$2.6} &  \textbf{\grh{83.9} \tiny{$\pm$1.9}} \\
\midrule
\textbf{Rank     } & &                  4.43 &                  \textbf{2.57} &                  7.14 &                  4.29 &                   4.71 &                   3.86 &                  4.29 &                  4.29 \\
\bottomrule
\end{tabular}
}
\label{tab:results}
\end{table}
Table~\ref{tab:results} reports the average and standard deviation of the classification accuracy on the test set, as well as the average ranking of the operators. We also report as baseline the classification accuracy of a GNN with no pooling (No-pool).
We observe that, on the datasets considered here, the operators based on graph spectral properties (MinCut, NDP, Graclus) achieve the highest accuracy.
However, we could not find strong evidence that one pooling operator (or even a class of operators) is systematically better than all others. 
For instance, on Triangles and Colors-3 we see that dense, trainable operators have a consistent advantage.
However, the family of sparse and/or non-trainable methods achieves a better performance on Enzymes, Mutagenicity, and the large-scale ModelNet10 dataset. Finally, in datasets such as Mutagenicity, Proteins, and DD the gap in performance is not very large.
Table~\ref{tab:results} also shows that some of the models with graph pooling operators achieve higher classification accuracy than the no-pool baseline (\grh{in green}). 
In Mutagenicity the baseline architecture achieves top performance, suggesting that graph pooling is not always beneficial in certain graph classification tasks. 
Further discussion can be found in the recent work of~\citet{mesquita2020rethinking}.

\section{Conclusions}
\label{sec:conclusions}
In this paper we presented SRC, a unifying formulation of pooling operators in GNNs, that allowed us to organize the vast literature on the subject under a comprehensive taxonomy and implement every pooling operator under a well-defined and modular framework.
Based on our framework, we gave pointers towards a possible implementation of a ``universal" pooling strategy based on well-known theory about the expressive power of GNNs. 
We identified three possible evaluation criteria for pooling operators and compared the different classes of the taxonomy on synthetic and real-world benchmarks. 
Results showed that pooling operators are not all equivalent and, in particular, there is no clear evidence that a particular pooling operator is consistently better than the others according to the evaluation criteria taken into consideration.
In our study, we showed the characteristics of different classes of operators and we reported an in-depth analysis of their inner mechanisms. 

Overall, we showed that the choice of the best pooling operator, and whether performing graph pooling is necessary at all, highly depends on the problem at hand. A comprehensive evaluation of pooling operators requires considering multiple criteria to highlight all their fundamental properties and, as such, it cannot be limited to measuring the downstream performance on few small-scale benchmark datasets.

\paragraph{Guidelines} 
For the above reasons, we provide guidelines to choose a pooling method in practice, based on the desiderata and the taxonomy:
\begin{itemize}
    \item To preserve node attributes, especially concerning geometric graphs, non-trainable and sparse methods are suggested since they usually compute a uniform coarsening of the graph;
    \item To preserve structure, dense and trainable methods are better at minimizing the Laplacian quadratic loss although sparse methods yield a better spectral alignment;
    \item To preserve task-specific information, which is the most common setting in machine learning, trainable methods have an edge over their counterpart although our conclusion is that there is no better method a priori.
    Non-trainable sparse methods (Graclus, NDP) have overall better performance across different tasks and are advised as the first choice; however, if the goal is to optimize for a specific objective, then trainable dense methods (MinCut, DiffPool) offer more flexibility and are more easily integrated into GNN architectures.
    \item We also observe that trainable sparse methods discard entire portions of the graphs and, therefore, they are generally less advisable.
\end{itemize}
We believe that SRC will be helpful in further studying graph pooling, and that our analysis will guide practitioners in choosing an appropriate pooling method for the application at hand. Our work also leads to a more principled approach in designing and evaluating new pooling operators, and will help the research community to advance the field. 

\section*{Acknowledgments}
This research is funded by the Swiss National Science Foundation project 200021 172671 ``ALPSFORT''. We gratefully acknowledge the support of Nvidia Corporation with the donation of the Titan XP GPU used for this work.

\bibliographystyle{abbrvnat}
\bibliography{references}

\begin{thebibliography}{59}
\providecommand{\natexlab}[1]{#1}
\providecommand{\url}[1]{\texttt{#1}}
\expandafter\ifx\csname urlstyle\endcsname\relax
  \providecommand{\doi}[1]{doi: #1}\else
  \providecommand{\doi}{doi: \begingroup \urlstyle{rm}\Url}\fi

\bibitem[Abadi et~al.(2016)Abadi, Agarwal, Barham, Brevdo, Chen, Citro,
  Corrado, Davis, Dean, Devin, et~al.]{abadi2016tensorflow}
M.~Abadi, A.~Agarwal, P.~Barham, E.~Brevdo, Z.~Chen, C.~Citro, G.~S. Corrado,
  A.~Davis, J.~Dean, M.~Devin, et~al.
\newblock Tensorflow: Large-scale machine learning on heterogeneous distributed
  systems.
\newblock \emph{arXiv preprint arXiv:1603.04467}, 2016.

\bibitem[Atwood and Towsley(2016)]{atwood2016diffusion}
J.~Atwood and D.~Towsley.
\newblock Diffusion-convolutional neural networks.
\newblock In \emph{Advances in Neural Information Processing Systems}, pages
  1993--2001, 2016.

\bibitem[Bacciu and Di~Sotto(2019)]{bacciu2019non}
D.~Bacciu and L.~Di~Sotto.
\newblock A non-negative factorization approach to node pooling in graph
  convolutional neural networks.
\newblock In \emph{Proceedings of the 18th International Conference of the
  Italian Association for Artificial Intelligence}. AIIA, 2019.

\bibitem[Bai et~al.(2019)Bai, Ding, Qiao, Marinovic, Gu, Chen, Sun, and
  Wang]{bai2019unsupervised}
Y.~Bai, H.~Ding, Y.~Qiao, A.~Marinovic, K.~Gu, T.~Chen, Y.~Sun, and W.~Wang.
\newblock Unsupervised inductive graph-level representation learning via
  graph-graph proximity.
\newblock \emph{arXiv preprint arXiv:1904.01098}, 2019.

\bibitem[Battaglia et~al.(2018)Battaglia, Hamrick, Bapst, Sanchez-Gonzalez,
  Zambaldi, Malinowski, Tacchetti, Raposo, Santoro, Faulkner,
  et~al.]{battaglia2018relational}
P.~W. Battaglia, J.~B. Hamrick, V.~Bapst, A.~Sanchez-Gonzalez, V.~Zambaldi,
  M.~Malinowski, A.~Tacchetti, D.~Raposo, A.~Santoro, R.~Faulkner, et~al.
\newblock Relational inductive biases, deep learning, and graph networks.
\newblock \emph{arXiv preprint arXiv:1806.01261}, 2018.

\bibitem[Bianchi et~al.(2020{\natexlab{a}})Bianchi, Grattarola, and
  Alippi]{bianchi2019mincut}
F.~M. Bianchi, D.~Grattarola, and C.~Alippi.
\newblock Spectral clustering with graph neural networks for graph pooling.
\newblock In \emph{International Conference on Machine Learning}, pages
  874--883. PMLR, 2020{\natexlab{a}}.

\bibitem[Bianchi et~al.(2020{\natexlab{b}})Bianchi, Grattarola, Livi, and
  Alippi]{bianchi2019hierarchical}
F.~M. Bianchi, D.~Grattarola, L.~Livi, and C.~Alippi.
\newblock Hierarchical representation learning in graph neural networks with
  node decimation pooling.
\newblock \emph{IEEE Transactions on Neural Networks and Learning Systems},
  2020{\natexlab{b}}.

\bibitem[Bodnar et~al.(2020)Bodnar, Cangea, and Li{\`o}]{bodnar2020deep}
C.~Bodnar, C.~Cangea, and P.~Li{\`o}.
\newblock Deep graph mapper: Seeing graphs through the neural lens.
\newblock \emph{arXiv preprint arXiv:2002.03864}, 2020.

\bibitem[Bruna et~al.(2013)Bruna, Zaremba, Szlam, and LeCun]{bruna2013spectral}
J.~Bruna, W.~Zaremba, A.~Szlam, and Y.~LeCun.
\newblock Spectral networks and locally connected networks on graphs.
\newblock \emph{arXiv preprint arXiv:1312.6203}, 2013.

\bibitem[Cai et~al.(2021)Cai, Wang, and Wang]{cai2021graph}
C.~Cai, D.~Wang, and Y.~Wang.
\newblock Graph coarsening with neural networks.
\newblock \emph{arXiv preprint arXiv:2102.01350}, 2021.

\bibitem[Cangea et~al.(2018)Cangea, Veli{\'{c}}kovi{\'c}, Jovanovi{\'c}, Kipf,
  and Li{\`o}]{cangea2018towards}
C.~Cangea, P.~Veli{\'{c}}kovi{\'c}, N.~Jovanovi{\'c}, T.~Kipf, and P.~Li{\`o}.
\newblock Towards sparse hierarchical graph classifiers.
\newblock \emph{arXiv preprint arXiv:1811.01287}, 2018.

\bibitem[Coates and Ng(2011)]{coates2011selecting}
A.~Coates and A.~Y. Ng.
\newblock Selecting receptive fields in deep networks.
\newblock In \emph{Advances in neural information processing systems}, pages
  2528--2536, 2011.

\bibitem[Corcoran(2019)]{corcoran2019function}
P.~Corcoran.
\newblock Function space pooling for graph convolutional networks.
\newblock \emph{arXiv preprint arXiv:1905.06259}, 2019.

\bibitem[Defferrard et~al.()Defferrard, Martin, Pena, and Perraudin]{pygsp}
M.~Defferrard, L.~Martin, R.~Pena, and N.~Perraudin.
\newblock Pygsp: Graph signal processing in python.
\newblock URL \url{https://github.com/epfl-lts2/pygsp/}.

\bibitem[Defferrard et~al.(2016)Defferrard, Bresson, and
  Vandergheynst]{defferrard2016convolutional}
M.~Defferrard, X.~Bresson, and P.~Vandergheynst.
\newblock Convolutional neural networks on graphs with fast localized spectral
  filtering.
\newblock In \emph{Advances in Neural Information Processing Systems}, pages
  3844--3852, 2016.

\bibitem[Dhillon et~al.(2007)Dhillon, Guan, and Kulis]{dhillon2007weighted}
I.~S. Dhillon, Y.~Guan, and B.~Kulis.
\newblock Weighted graph cuts without eigenvectors a multilevel approach.
\newblock \emph{IEEE transactions on pattern analysis and machine
  intelligence}, 29\penalty0 (11):\penalty0 1944--1957, 2007.

\bibitem[Diehl(2019)]{diehl2019edge}
F.~Diehl.
\newblock Edge contraction pooling for graph neural networks.
\newblock \emph{CoRR}, abs/1905.10990, 2019.
\newblock URL \url{http://arxiv.org/abs/1905.10990}.

\bibitem[{Dorfler} and {Bullo}(2013)]{kron_red}
F.~{Dorfler} and F.~{Bullo}.
\newblock Kron reduction of graphs with applications to electrical networks.
\newblock \emph{IEEE Transactions on Circuits and Systems I: Regular Papers},
  60\penalty0 (1):\penalty0 150--163, Jan 2013.
\newblock ISSN 1549-8328.
\newblock \doi{10.1109/TCSI.2012.2215780}.

\bibitem[Fey and Lenssen(2019)]{fey2019fast}
M.~Fey and J.~E. Lenssen.
\newblock Fast graph representation learning with pytorch geometric.
\newblock \emph{arXiv preprint arXiv:1903.02428}, 2019.

\bibitem[Fey et~al.(2018)Fey, Lenssen, Weichert, and
  M{\"u}ller]{fey2018splinecnn}
M.~Fey, J.~E. Lenssen, F.~Weichert, and H.~M{\"u}ller.
\newblock Splinecnn: Fast geometric deep learning with continuous b-spline
  kernels.
\newblock In \emph{Proceedings of the IEEE Conference on Computer Vision and
  Pattern Recognition}, pages 869--877, 2018.

\bibitem[Gilmer et~al.(2017)Gilmer, Schoenholz, Riley, Vinyals, and
  Dahl]{gilmer2017neural}
J.~Gilmer, S.~S. Schoenholz, P.~F. Riley, O.~Vinyals, and G.~E. Dahl.
\newblock Neural message passing for quantum chemistry.
\newblock \emph{arXiv preprint arXiv:1704.01212}, 2017.

\bibitem[Grattarola and Alippi(2021)]{grattarola2020graph}
D.~Grattarola and C.~Alippi.
\newblock Graph neural networks in tensorflow and keras with spektral.
\newblock \emph{IEEE Computational Intelligence Magazine}, 2021.

\bibitem[Hermsdorff and Gunderson(2019)]{hermsdorff2019unifying}
G.~B. Hermsdorff and L.~M. Gunderson.
\newblock A unifying framework for spectrum-preserving graph sparsification and
  coarsening.
\newblock \emph{arXiv preprint arXiv:1902.09702}, 2019.

\bibitem[Hongyang~Gao(2019)]{graphunet}
S.~J. Hongyang~Gao.
\newblock Graph u-net.
\newblock \emph{Submitted to ICLR}, 2019.

\bibitem[Ioffe and Szegedy(2015)]{ioffe2015batch}
S.~Ioffe and C.~Szegedy.
\newblock Batch normalization: Accelerating deep network training by reducing
  internal covariate shift.
\newblock In \emph{International conference on machine learning}, pages
  448--456. PMLR, 2015.

\bibitem[Karypis(1997)]{karypis1997metis}
G.~Karypis.
\newblock Metis: Unstructured graph partitioning and sparse matrix ordering
  system.
\newblock \emph{Technical report}, 1997.

\bibitem[Keriven and Peyr\'{e}(2019)]{keriven2019universal}
N.~Keriven and G.~Peyr\'{e}.
\newblock Universal invariant and equivariant graph neural networks.
\newblock In \emph{Advances in Neural Information Processing Systems 32}, pages
  7090--7099. Curran Associates, Inc., 2019.
\newblock URL
  \url{http://papers.nips.cc/paper/8931-universal-invariant-and-equivariant-graph-neural-networks.pdf}.

\bibitem[Knyazev et~al.(2019)Knyazev, Taylor, and
  Amer]{DBLP:journals/corr/abs-1905-02850}
B.~Knyazev, G.~W. Taylor, and M.~R. Amer.
\newblock Understanding attention in graph neural networks.
\newblock \emph{CoRR}, abs/1905.02850, 2019.
\newblock URL \url{http://arxiv.org/abs/1905.02850}.

\bibitem[Kushnir et~al.(2006)Kushnir, Galun, and Brandt]{kushnir2006fast}
D.~Kushnir, M.~Galun, and A.~Brandt.
\newblock Fast multiscale clustering and manifold identification.
\newblock \emph{Pattern Recognition}, 39\penalty0 (10):\penalty0 1876--1891,
  2006.

\bibitem[Lee et~al.(2019)Lee, Lee, and Kang]{lee2019self}
J.~Lee, I.~Lee, and J.~Kang.
\newblock Self-attention graph pooling.
\newblock \emph{arXiv preprint arXiv:1904.08082}, 2019.

\bibitem[Lei et~al.(2019)Lei, Akhtar, and Mian]{lei2019octree}
H.~Lei, N.~Akhtar, and A.~Mian.
\newblock Octree guided cnn with spherical kernels for 3d point clouds.
\newblock In \emph{Proceedings of the IEEE Conference on Computer Vision and
  Pattern Recognition}, pages 9631--9640, 2019.

\bibitem[Levie et~al.(2017)Levie, Monti, Bresson, and
  Bronstein]{levie2017cayleynets}
R.~Levie, F.~Monti, X.~Bresson, and M.~M. Bronstein.
\newblock Cayleynets: Graph convolutional neural networks with complex rational
  spectral filters.
\newblock \emph{arXiv preprint arXiv:1705.07664}, 2017.

\bibitem[Li et~al.(2015)Li, Tarlow, Brockschmidt, and Zemel]{li2015gated}
Y.~Li, D.~Tarlow, M.~Brockschmidt, and R.~Zemel.
\newblock Gated graph sequence neural networks.
\newblock \emph{arXiv preprint arXiv:1511.05493}, 2015.

\bibitem[Loukas(2019)]{loukas2019graph}
A.~Loukas.
\newblock Graph reduction with spectral and cut guarantees.
\newblock \emph{Journal of Machine Learning Research}, 20\penalty0
  (116):\penalty0 1--42, 2019.

\bibitem[Luzhnica et~al.(2019)Luzhnica, Day, and Lio]{luzhnica2019clique}
E.~Luzhnica, B.~Day, and P.~Lio.
\newblock Clique pooling for graph classification.
\newblock \emph{International Conference of Learning Representations (ICLR) --
  Representation Learning on Graphs and Manifolds workshop}, 2019.

\bibitem[Ma et~al.(2019)Ma, Wang, Aggarwal, and Tang]{ma2019graph}
Y.~Ma, S.~Wang, C.~C. Aggarwal, and J.~Tang.
\newblock Graph convolutional networks with eigenpooling.
\newblock \emph{arXiv preprint arXiv:1904.13107}, 2019.

\bibitem[Maron et~al.(2019)Maron, Fetaya, Segol, and
  Lipman]{maron2019universality}
H.~Maron, E.~Fetaya, N.~Segol, and Y.~Lipman.
\newblock On the universality of invariant networks.
\newblock In \emph{International Conference on Machine Learning}, pages
  4363--4371, 2019.

\bibitem[Mesquita et~al.(2020)Mesquita, Souza, and
  Kaski]{mesquita2020rethinking}
D.~Mesquita, A.~Souza, and S.~Kaski.
\newblock Rethinking pooling in graph neural networks.
\newblock \emph{Advances in Neural Information Processing Systems}, 33, 2020.

\bibitem[Monti et~al.(2017)Monti, Boscaini, Masci, Rodola, Svoboda, and
  Bronstein]{monti2017geometric}
F.~Monti, D.~Boscaini, J.~Masci, E.~Rodola, J.~Svoboda, and M.~M. Bronstein.
\newblock Geometric deep learning on graphs and manifolds using mixture model
  cnns.
\newblock In \emph{Proc. CVPR}, volume~1, page~3, 2017.

\bibitem[Morris et~al.(2020)Morris, Kriege, Bause, Kersting, Mutzel, and
  Neumann]{morris2020tudataset}
C.~Morris, N.~M. Kriege, F.~Bause, K.~Kersting, P.~Mutzel, and M.~Neumann.
\newblock Tudataset: A collection of benchmark datasets for learning with
  graphs.
\newblock \emph{arXiv preprint arXiv:2007.08663}, 2020.

\bibitem[Navarin et~al.(2019)Navarin, Van~Tran, and
  Sperduti]{navarin2019universal}
N.~Navarin, D.~Van~Tran, and A.~Sperduti.
\newblock Universal readout for graph convolutional neural networks.
\newblock In \emph{2019 International Joint Conference on Neural Networks
  (IJCNN)}, pages 1--7. IEEE, 2019.

\bibitem[Noutahi et~al.(2019)Noutahi, Beani, Horwood, and
  Tossou]{noutahi2019towards}
E.~Noutahi, D.~Beani, J.~Horwood, and P.~Tossou.
\newblock Towards interpretable sparse graph representation learning with
  laplacian pooling.
\newblock \emph{arXiv preprint arXiv:1905.11577}, 2019.

\bibitem[Qi et~al.(2017)Qi, Yi, Su, and Guibas]{qi2017pointnet++}
C.~R. Qi, L.~Yi, H.~Su, and L.~J. Guibas.
\newblock Pointnet++: Deep hierarchical feature learning on point sets in a
  metric space.
\newblock In \emph{Advances in neural information processing systems}, pages
  5099--5108, 2017.

\bibitem[Ranjan et~al.(2019)Ranjan, Sanyal, and Talukdar]{ranjan2019asap}
E.~Ranjan, S.~Sanyal, and P.~P. Talukdar.
\newblock Asap: Adaptive structure aware pooling for learning hierarchical
  graph representations.
\newblock \emph{arXiv preprint arXiv:1911.07979}, 2019.

\bibitem[Riegler et~al.(2017)Riegler, Osman~Ulusoy, and
  Geiger]{riegler2017octnet}
G.~Riegler, A.~Osman~Ulusoy, and A.~Geiger.
\newblock Octnet: Learning deep 3d representations at high resolutions.
\newblock In \emph{Proceedings of the IEEE Conference on Computer Vision and
  Pattern Recognition}, pages 3577--3586, 2017.

\bibitem[Sato(2020)]{sato2020survey}
R.~Sato.
\newblock A survey on the expressive power of graph neural networks.
\newblock \emph{arXiv preprint arXiv:2003.04078}, 2020.

\bibitem[Scarselli et~al.(2009)Scarselli, Gori, Tsoi, Hagenbuchner, and
  Monfardini]{scarselli2009graph}
F.~Scarselli, M.~Gori, A.~C. Tsoi, M.~Hagenbuchner, and G.~Monfardini.
\newblock The graph neural network model.
\newblock \emph{IEEE Transactions on Neural Networks}, 20\penalty0
  (1):\penalty0 61--80, 2009.

\bibitem[Simonovsky and Komodakis(2017)]{simonovsky2017dynamic}
M.~Simonovsky and N.~Komodakis.
\newblock Dynamic edgeconditioned filters in convolutional neural networks on
  graphs.
\newblock In \emph{Proc. CVPR}, 2017.

\bibitem[Vinyals et~al.(2015)Vinyals, Bengio, and Kudlur]{vinyals2015order}
O.~Vinyals, S.~Bengio, and M.~Kudlur.
\newblock Order matters: Sequence to sequence for sets.
\newblock \emph{arXiv preprint arXiv:1511.06391}, 2015.

\bibitem[Von~Luxburg(2007)]{von2007tutorial}
U.~Von~Luxburg.
\newblock A tutorial on spectral clustering.
\newblock \emph{Statistics and computing}, 17\penalty0 (4):\penalty0 395--416,
  2007.

\bibitem[Wu et~al.(2019)Wu, He, and Xu]{wu2019net}
J.~Wu, J.~He, and J.~Xu.
\newblock Net: Degree-specific graph neural networks for node and graph
  classification.
\newblock \emph{arXiv preprint arXiv:1906.02319}, 2019.

\bibitem[Wu et~al.(2015)Wu, Song, Khosla, Yu, Zhang, Tang, and Xiao]{wu20153d}
Z.~Wu, S.~Song, A.~Khosla, F.~Yu, L.~Zhang, X.~Tang, and J.~Xiao.
\newblock 3d shapenets: A deep representation for volumetric shapes.
\newblock In \emph{Proceedings of the IEEE conference on computer vision and
  pattern recognition}, pages 1912--1920, 2015.

\bibitem[Xie et~al.(2020)Xie, Yao, Gong, Chen, and Qin]{xie2020graph}
Y.~Xie, C.~Yao, M.~Gong, C.~Chen, and A.~Qin.
\newblock Graph convolutional networks with multi-level coarsening for graph
  classification.
\newblock \emph{Knowledge-Based Systems}, page 105578, 2020.

\bibitem[Xu et~al.(2018)Xu, Hu, Leskovec, and Jegelka]{xu2018powerful}
K.~Xu, W.~Hu, J.~Leskovec, and S.~Jegelka.
\newblock How powerful are graph neural networks?
\newblock \emph{arXiv preprint arXiv:1810.00826}, 2018.

\bibitem[Ying et~al.(2018)Ying, You, Morris, Ren, Hamilton, and
  Leskovec]{ying2018hierarchical}
R.~Ying, J.~You, C.~Morris, X.~Ren, W.~L. Hamilton, and J.~Leskovec.
\newblock Hierarchical graph representation learning with differentiable
  pooling.
\newblock \emph{arXiv preprint arXiv:1806.08804}, 2018.

\bibitem[You et~al.(2020)You, Ying, and Leskovec]{you2020design}
J.~You, Z.~Ying, and J.~Leskovec.
\newblock Design space for graph neural networks.
\newblock \emph{Advances in Neural Information Processing Systems}, 33, 2020.

\bibitem[Yuan and Ji(2020)]{yuan2020structpool}
H.~Yuan and S.~Ji.
\newblock Structpool: Structured graph pooling via conditional random fields.
\newblock In \emph{International Conference on Learning Representations}, 2020.
\newblock URL \url{https://openreview.net/forum?id=BJxg_hVtwH}.

\bibitem[Zhang et~al.(2018)Zhang, Cui, Neumann, and Chen]{zhang2018end}
M.~Zhang, Z.~Cui, M.~Neumann, and Y.~Chen.
\newblock An end-to-end deep learning architecture for graph classification.
\newblock In \emph{Proceedings of AAAI Conference on Artificial Inteligence},
  2018.

\bibitem[Zhang et~al.(2021)Zhang, Bu, Ester, Zhang, Li, Yao, Huifen, Yu, and
  Wang]{zhang2021hierarchical}
Z.~Zhang, J.~Bu, M.~Ester, J.~Zhang, Z.~Li, C.~Yao, D.~Huifen, Z.~Yu, and
  C.~Wang.
\newblock Hierarchical multi-view graph pooling with structure learning.
\newblock \emph{IEEE Transactions on Knowledge and Data Engineering}, 2021.

\end{thebibliography}

\appendix
\section{Experimental Details}
\label{app:experiments}

\def\GNN{\textsc{GNN}}
\def\MLP{\textsc{MLP}}
\def\Pool{\textsc{Pool}}

\subsection{Hardware and software}
All experiments were run on an NVIDIA Titan V GPU with 12 GB of video memory. 
All models were implemented in Python 3.7, and all machine learning code was based on TensorFlow 2.4 \cite{abadi2016tensorflow} and Spektral 1.0.6 \cite{grattarola2020graph}.
The code to reproduce the experiments is available at \codeurl.

\subsection{Preliminaries} 
We use the same message-passing model in all of our experiments. The model is inspired by the work of \citet{you2020design} and has the following form: 
\begin{equation}
\label{eq:gnn}
    \x_i' = \x_i \;\Big\| \sum\limits_{j \in \mathcal{N}(i) } \mathrm{ReLU} \left( \mathrm{BN} \left( \W\x_j + \b \right) \right)
\end{equation}
where $\mathrm{BN}$ indicates batch normalization \cite{ioffe2015batch}, $\mathrm{ReLU}$ is the rectified linear unit, $\x_i \in \mathbb{R}^{d_{in}}$, $\x_i' \in \mathbb{R}^{d_{out}}$, $\W \in \mathbb{R}^{d_{out} \times d_{in}}$ is a trainable matrix, $\b \in \mathbb{R}^{d_{out}}$ is a trainable bias, and $\Big\|$ indicates concatenation. We configure all layers to have $d_{out} = 256$.

We use $\GNN(\A, \X)$ to indicate the application of one layer as in Eq.\ \eqref{eq:gnn} to a graph described by $\A$ and $\X$.

We use $\MLP(\X)$ to indicate the application, to the features of each node, of one multi-layer perceptron (MLP) with 2 layers, 256 hidden units, ReLU activation and batch normalization. The number of neurons of each layer is implied by the dimension of the data or, if not specified, is also set to 256 units (\eg, for MLPs used as intermediate blocks).

All architectures and hyperparameters are also inspired by the work \citet{you2020design}. Specifically, we include a pre-processing and post-processing MLP as first and last blocks of each architecture, respectively.
The activation of the last layer of the post-processing MLP is implied by the task.

We use $\Pool(\A, \X)$ to indicate the application of a generic pooling layer, which is substituted in the architecture with the different operators that we compared in our experiments. Note that, for clarity, we change notation w.r.t.\ the main text and indicate with $\X_{pool}$, $\A_{pool}$ the output of a pooling layer.

\subsection{Preserving node attributes}
\paragraph{Architecture} The bottleneck of the autoencoder consists of a single pooling layer to compress the graph signal, immediately followed by an \emph{upscaling} layer. The overall architecture is:
\begin{align*}
    \X, \A &\leftarrow \cG\\
    \X_{in} &\leftarrow \MLP_{in}\left(\X\right) \\ 
    \X_{in} &\leftarrow \GNN_{in}\left(\A, \X_{in}\right) \\ 
    \X_{pool} &\leftarrow \Pool\left(\A, \X_{in}\right) \\
    \X_{up} &\leftarrow \UpScale\left(\X_{pool}\right) \\
    \X_{out} &\leftarrow \GNN_{out}\left(\A, \X_{up}\right)\\
    \X_{out} &\leftarrow \MLP_{out}\left(\X_{out}\right).
\end{align*}
All pooling layers are configured to reduce the graph to $K = \left\lfloor N/2 \right\rfloor$ nodes, except for LaPool which determines the number of output nodes autonomously.

The $\UpScale$ layer lifts the reduced node features $\X'$ of the coarsened graph $\cG'$ back to the original data dimensionality of $\X_{in}$.
For almost all methods, the upscaled node features $\X_{up}$ are constructed as $\U \X'$, where $\U = \S^\mp$ is the transposed pseudo-inverse of the selection matrix $\S = \Sel(\X,\A)$. This is the optimal choice when the reduction function is $\X' = \S^\top \X_{in}$. Conversely, for Top-$K$ and SAGPool we need to account also for the scaling factors $\sigma(\y)$, so we have $\U=(\sigma(\y)\odot\S)^\mp$. Finally, even though DiffPool's reduction is $\X'=\S^\top \GNN(\A, \X)$, we still employ $\U\S^\mp$ as upscaling operator and allow the output layer $\GNN_{out}$ to counteract the effects of the $\GNN$.

Notice that $\GNN_{out}$ takes as input the original adjacency matrix $\A$, so as to focus the experiment on the node features only
and suppressing possible interference of the \Con\ function.

\paragraph{Training} All models are trained to convergence using Adam to minimize the mean squared error between the input and output features, with learning rate $0.0005$ and early stopping on the training loss with a patience of $1000$ epochs and a tolerance of $10^{-6}$. Each experiment is repeated 3 times.

\begin{table}[]
    \centering
    \caption{Statistics of graphs used in the autoencoder and spectral similarity experiments.}
    \label{tab:ae_details}
    \begin{tabular}{@{}llll@{}}
        \toprule
        \textbf{Graph} & \textbf{Nodes} & \textbf{Edges} & \textbf{Avg. degree} \\
        \midrule
        Grid & 64 & 112 & 3.5 \\
        Ring & 64 & 64 & 2 \\
        Bunny & 2503 & 65490 & 52.35 \\
        Airfoil & 4253 & 12289 & 5.77 \\
        Sensor & 64 & 313.7 $\pm$ 21.9 & 9.8 $\pm$ 0.8 \\
        \midrule
        Airplane & 1333 & 2611 & 3.91 \\
        Car & 1920 & 2372 & 2.47 \\
        Guitar & 3125 & 5508 & 3.52 \\
        Person & 3305 & 9055 & 5.47 \\
        \bottomrule
    \end{tabular}
\end{table}
\paragraph{Data} The Grid, Ring, and Bunny graphs are generated using the PyGSP library \cite{pygsp}. The Airplane, Car, Person, and Guitar graphs are taken from the ModelNet40 dataset \cite{wu20153d}. We selected one graph randomly from the training set of each category (with a threshold on the number of nodes). The ModelNet40 IDs of the selected graph are: sample 151 for Airplane, sample 75 for Car, sample 38 for Guitar, sample 83 for Person.
Details on the size and average degrees of the graphs are reported in Table \ref{tab:ae_details}.
All datasets that were not generated programmatically are unlicensed.

\subsection{Preserving structure}
\paragraph{Architecture} The architecture for this experiment consists only of a pooling layer, to ensure that the model is actually operating on the original coordinates and eigenvectors without transformations:
\begin{align*}
    \X, \A &\leftarrow \cG\\
    \X_{pool} &\leftarrow \Pool\left(\A, \X\right) \\
\end{align*}
All pooling layers are configured to reduce the graph to $K = \left\lfloor N/2 \right\rfloor$ nodes, except for LaPool which determines the number of output nodes autonomously.

\paragraph{Training} For trainable models, we train them to convergence using Adam to minimize the quadratic loss described in the main text, with learning rate $0.01$ and early stopping on the training loss with a patience of $50$ epochs and a tolerance of $10^{-6}$. Each experiment is repeated 3 times.

\paragraph{Data} All graphs are generated using the PyGSP library. In particular, Sensor is a random graph which is generated once per experiment and used for all models.
Details on the size and average degrees of the graphs are reported in Table \ref{tab:ae_details}. All datasets that were not generated programmatically are unlicensed.

\subsection{Preserving task-specific information}
\paragraph{Architecture} We use the following architecture for all datasets:
\begin{align*}
    \X, \A &\leftarrow \cG\\
    \X &\leftarrow \MLP_{1}\left(\X\right) \\ 
    \X &\leftarrow \GNN_{1}\left(\A, \X\right) \\ 
    \A_{pool}, \X_{pool} &\leftarrow \Pool\left(\A, \X\right) \\
    \X_{pool} &\leftarrow \GNN_{2}\left(\A_{pool},\X_{pool}\right) \\ 
    \x_{out} &\leftarrow \sum_{i} \x_{pool, i} \\
    \x_{out} &\leftarrow \MLP_{2}\left(\x_{out}\right) \\ 
\end{align*}
Adaptive pooling layers are configured to reduce the graph to $K = \left\lfloor N/2 \right\rfloor$ nodes, except for LaPool which determines the number of output nodes autonomously.
Fixed pooling layers are configured to return $K = \left\lfloor \bar N/2 \right\rfloor$ nodes, where $\bar N$ is the average number of nodes in the training set. 

\paragraph{Training} All models are trained to convergence using Adam, with a batch size of $16$, learning rate $0.0005$ and early stopping on the validation loss with a patience of $50$ epochs. Each experiment is repeated 3 times.

\paragraph{Data} Proteins, Enzymes, Mutagenicity, DD, Colors-3 and Triangles are taken from the TUDataset collection \cite{morris2020tudataset}. The ModelNet10 dataset is taken from its original source \cite{wu20153d}.
All datasets are split randomly according to an 8:1:1 proportion between training, validation and test sets. The only exceptions are Colors-3 and Triangles, for which the data splits are described in \cite{DBLP:journals/corr/abs-1905-02850}, and ModelNet10, for which the data splits are given.
The TUDatasets are unlicensed and ModelNet10 is provided freely for academic use.

\subsection{Memory usage}

To produce Figure 3, we generated random Erd\H{o}s-R\'enyi graphs with $p=0.1$ and random features, and gave it as input to the trainable pooling methods (MinCutPool, DiffPool, Top-$K$ and SAGPool). 
At each forward pass, we increased the number of nodes by 1000 until an out-of-memory exception was raised. 
We used a sparse tensor to represent the adjacency matrix of the input graphs (so that the cost of loading $\A$ into memory is linear in the number of edges).
We repeated the experiments with node features of size $F=1, 10, 100, 1000$ and found no significant differences in the results.

\section{Additional results}

\subsection{Preserving node attributes}

\begin{figure}
    \centering
    \includegraphics[width=\textwidth]{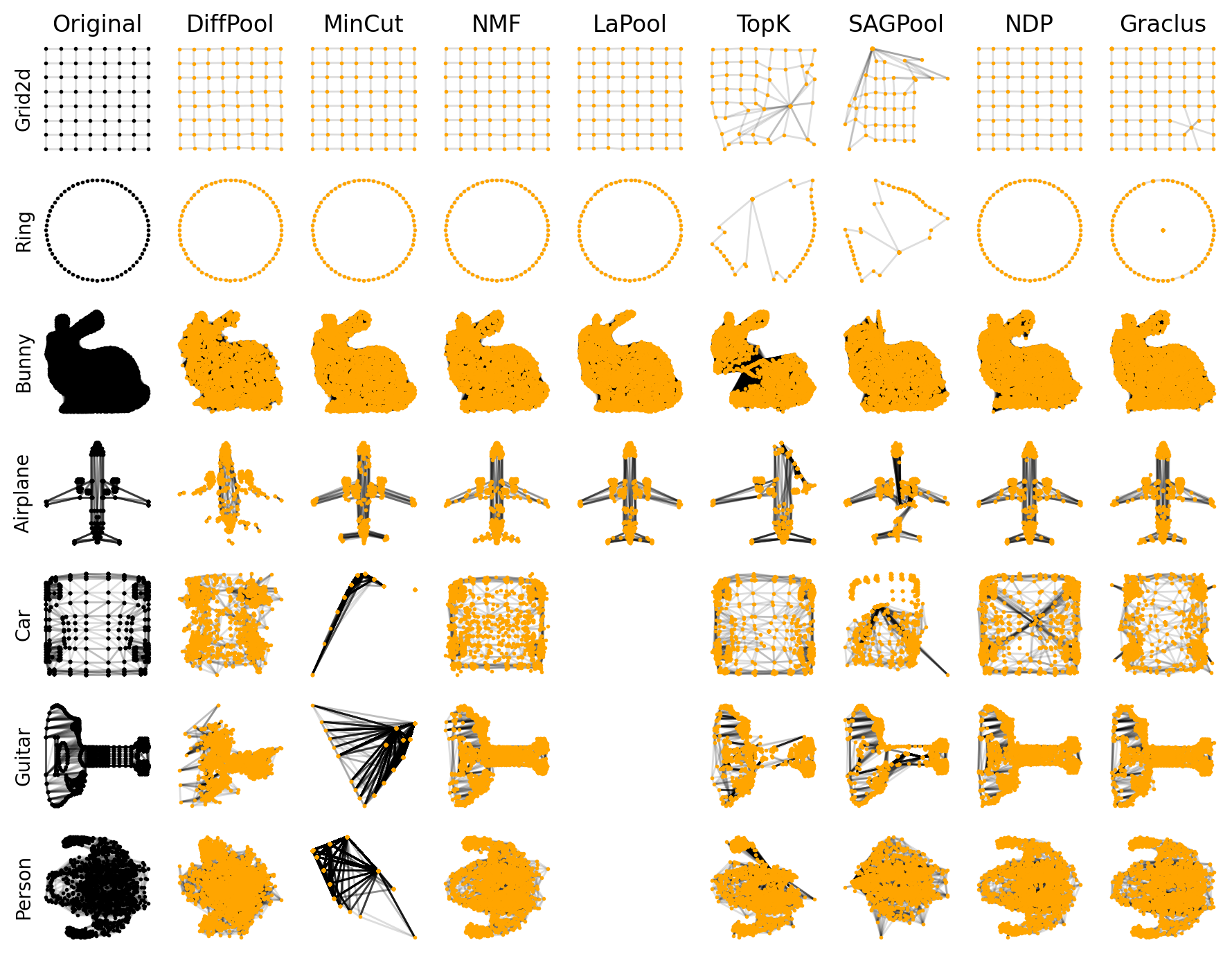}
    \caption{Node attributes (point locations) reconstructed with different operators in the autoencoder experiment.}
    \label{fig:ae_rec_graphs_all}
\end{figure}

\begin{figure}
    \centering
    \includegraphics[width=\textwidth]{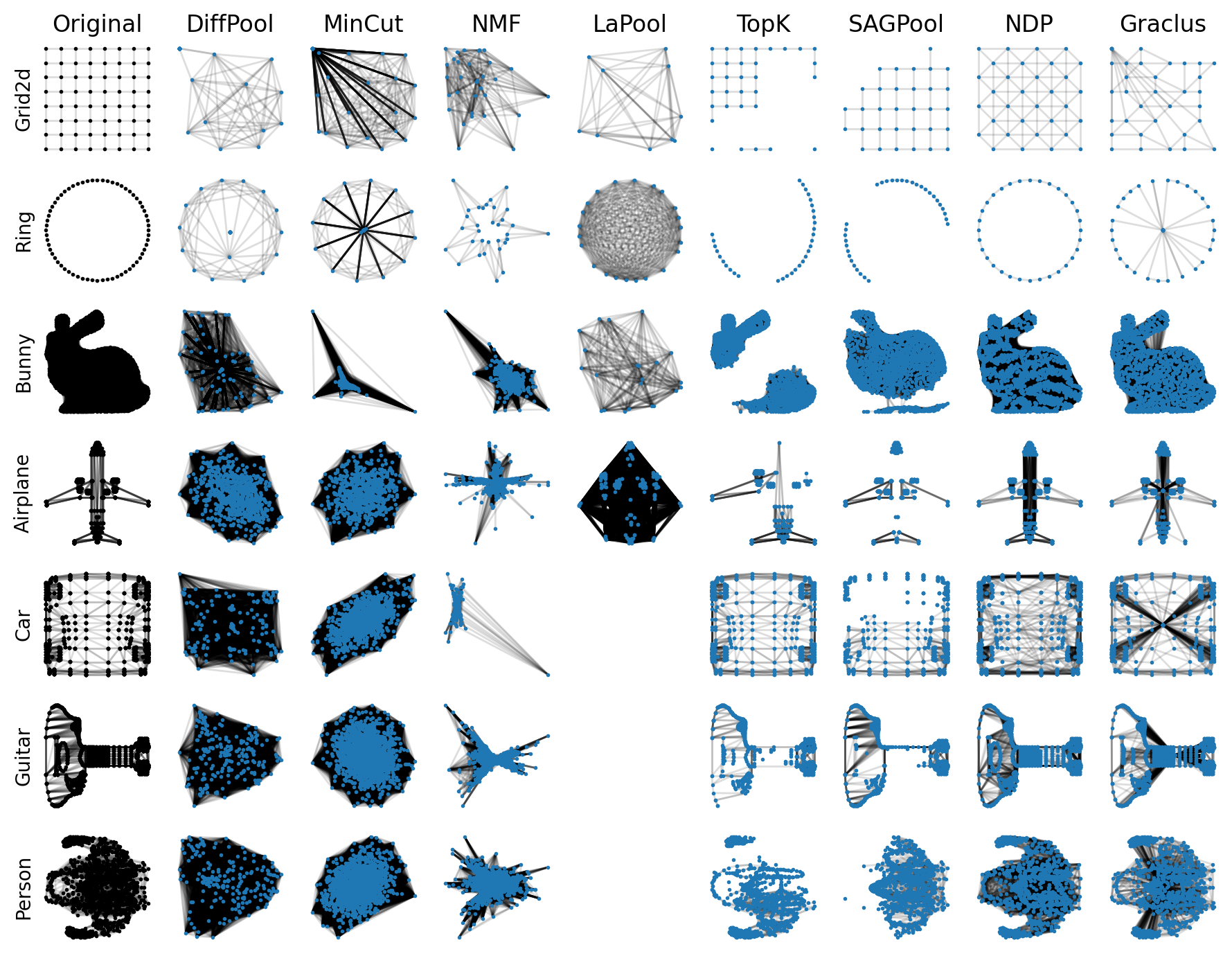}
    \caption{Graphs pooled with different operators in the autoencoder experiment with the modified \Red\ function.}
    \label{fig:ae_mod_red_all}
\end{figure}

\begin{figure}
    \centering
    \includegraphics[width=\textwidth]{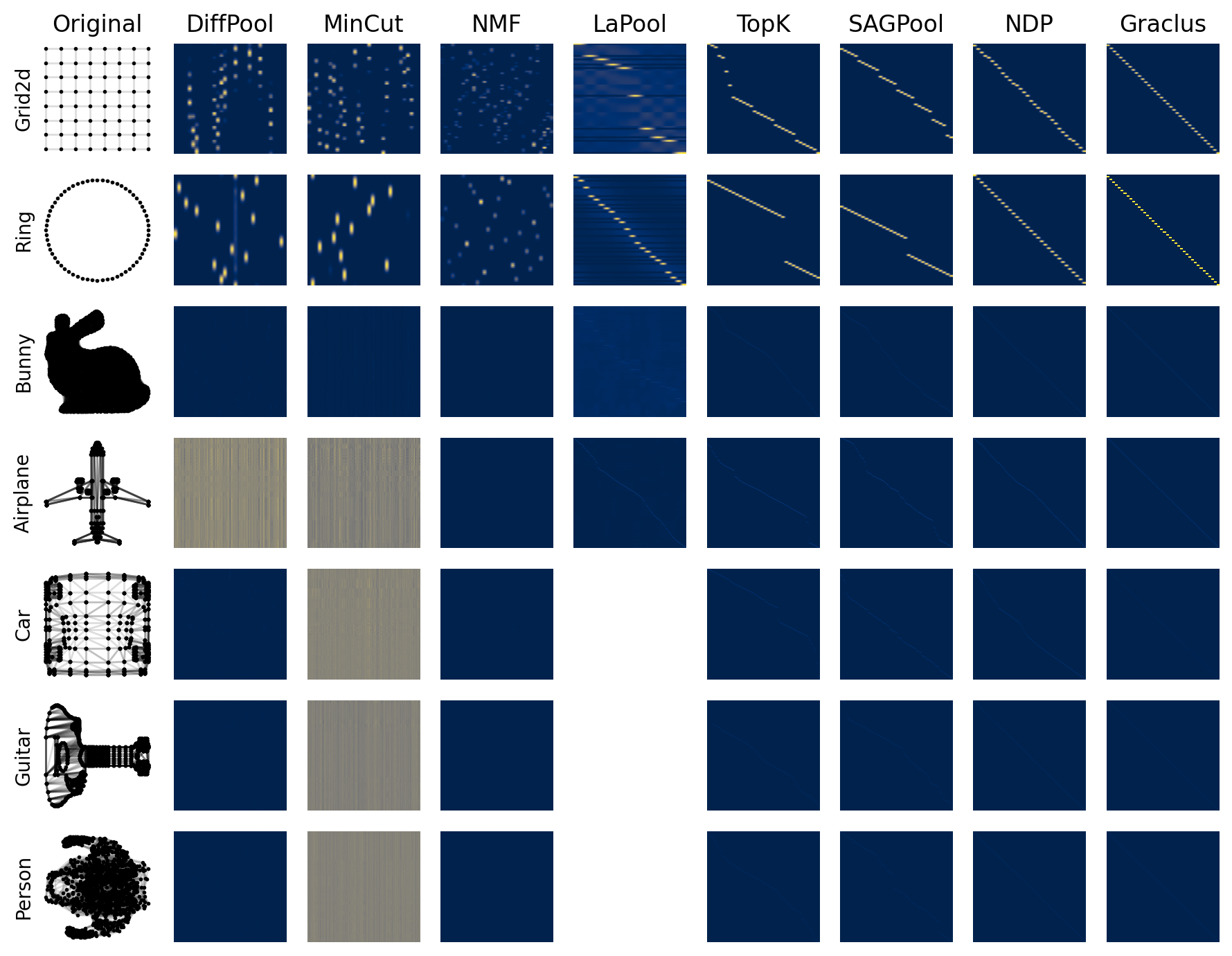}
    \caption{Selection matrices computed with different operators in the autoencoder experiment.}
    \label{fig:ae_sel_mat_all}
\end{figure}

An extended version of Figure 4 in the main paper is reported in Figure \ref{fig:ae_rec_graphs_all} here.
An extended version of Figure 5 in the main paper is reported in Figures \ref{fig:ae_mod_red_all} and \ref{fig:ae_sel_mat_all}. Note that the missing plots for LaPool are due to the Out Of Memory exception as reported in Table 3 in the main paper. 

\subsection{Preserving structure}

An extended version of Figure 6 in the main paper is reported in Figure \ref{fig:spectral_sim_all} here.
An extended version of Table 5 is reported in Table \ref{tab:ss_density_all}.

\begin{figure}
    \centering
    \subfigure[Grid]{\includegraphics[width=\textwidth]{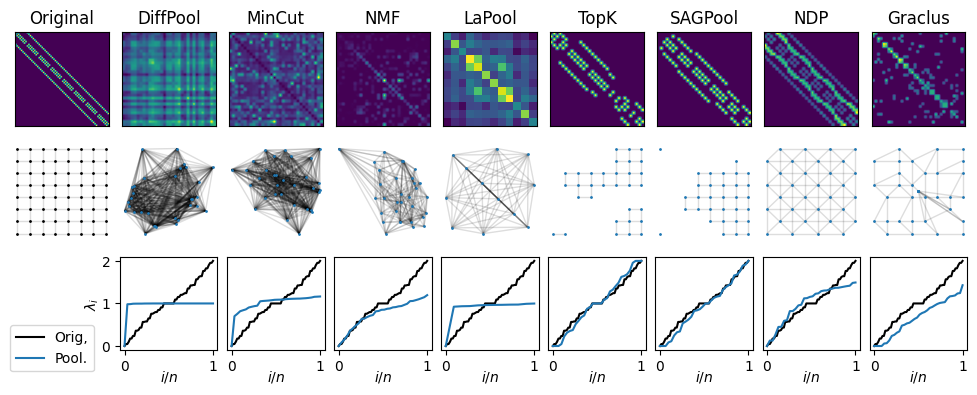}}
    \subfigure[Ring]{\includegraphics[width=\textwidth]{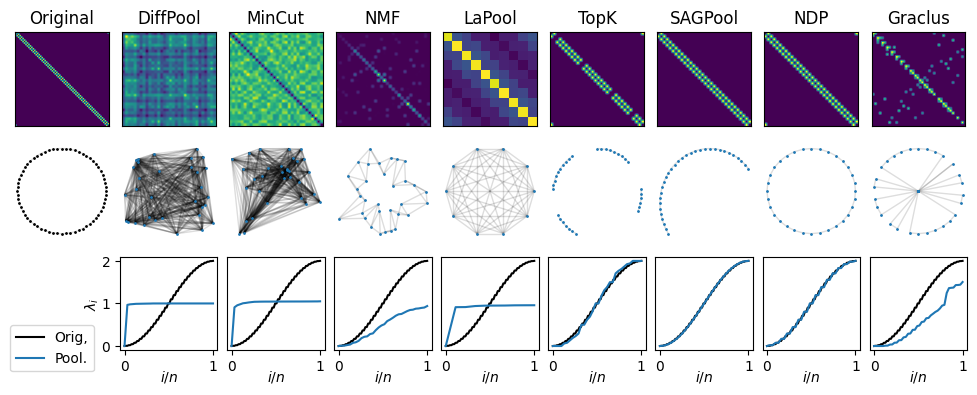}}
    \subfigure[Sensor]{\includegraphics[width=\textwidth]{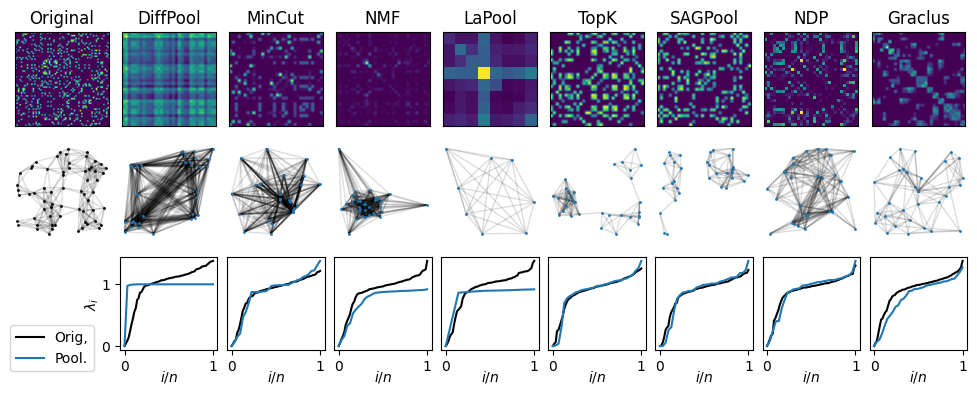}}
\end{figure}
\begin{figure}
    \centering
    \subfigure[Bunny]{\includegraphics[width=\textwidth]{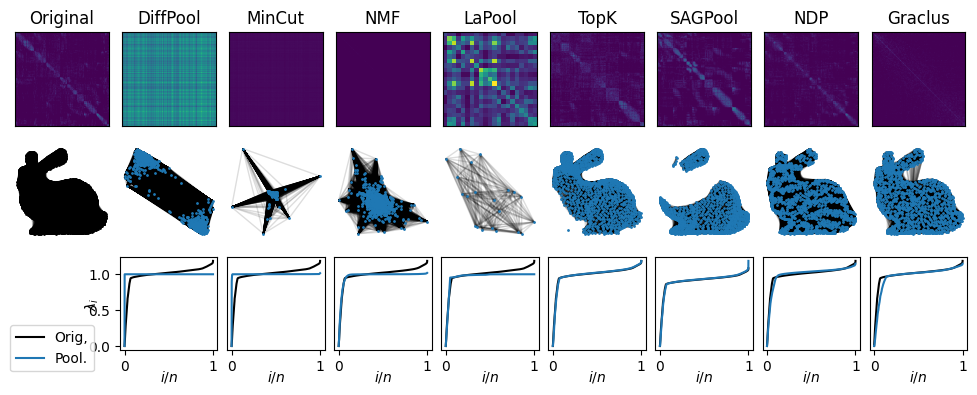}}
    \subfigure[Minnesota]{\includegraphics[width=\textwidth]{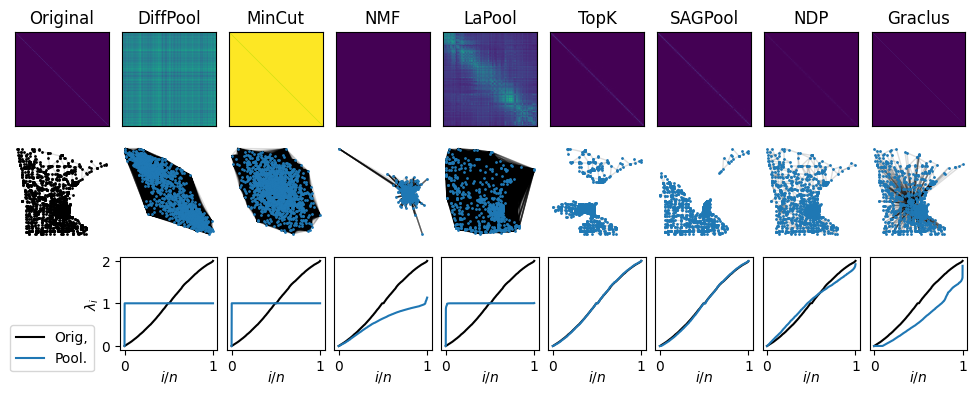}}
    \subfigure[Airfoil]{\includegraphics[width=\textwidth]{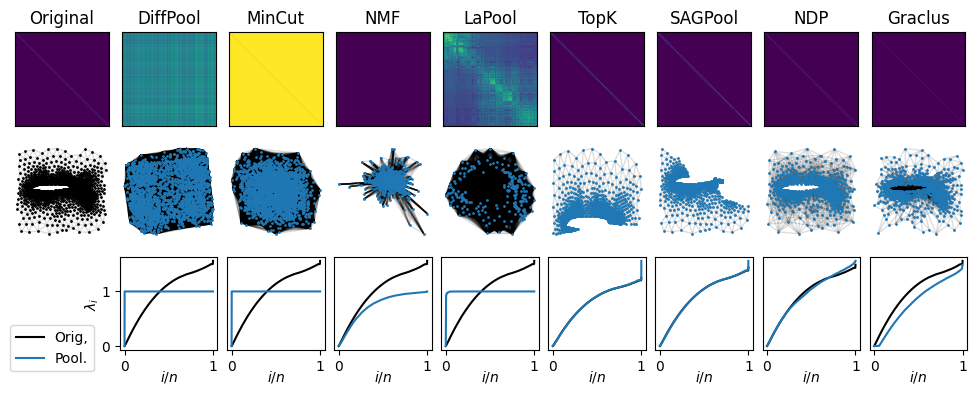}}
    \caption{\small Results for all graphs when optimizing for spectral similarity. Top: the coarsened adjacency matrices. Middle: the coarsened graphs with modified \Sel\ function. Bottom: the eigenvalues of the normalized Laplacian before (black) and after (blue) pooling. The indices of the eigenvalues are rescaled to fill $[0,1]$.}
    \label{fig:spectral_sim_all}
\end{figure}

\begin{table}[]
    \centering
    \caption{Density and median weight of the edges of the coarsened graphs in the spectral similarity experiment.}
    \label{tab:ss_density_all}
    \resizebox{\textwidth}{!}{%
    \begin{tabular}{@{}lllllllllll@{}}
    \toprule
            &        &    \textbf{Original} & \textbf{DiffPool} &      \textbf{MinCut} &         \textbf{NMF} & \textbf{LaPool} & \textbf{TopK} & \textbf{SAGPool} & \textbf{NDP} & \textbf{Graclus} \\
    \midrule
    \multirow{2}{*}{\textbf{Grid2d}} & \textbf{Density} &                0.055 &             0.969 &                0.969 &                0.463 &           0.917 &         0.084 &            0.092 &        0.189 &            0.103 \\
            & \textbf{Median} &                1.000 &             0.216 &                0.024 &                0.018 &           1.445 &         1.000 &            1.000 &        0.500 &            0.154 \\
    \midrule
    \multirow{2}{*}{\textbf{Ring}} & \textbf{Density} &                0.031 &             0.969 &                0.969 &                0.125 &           0.900 &         0.055 &            0.061 &        0.062 &            0.045 \\
            & \textbf{Median} &                1.000 &             0.124 &                0.032 &                0.039 &           1.171 &         1.000 &            1.000 &        0.500 &            0.250 \\
    \midrule
    \multirow{2}{*}{\textbf{Bunny}} & \textbf{Density} &                0.021 &             0.999 &                0.999 &                0.327 &           0.952 &         0.038 &            0.038 &        0.104 &            0.029 \\
            & \textbf{Median} &                0.812 &             0.068 &  7.55$\cdot 10^{-4}$ &  9.99$\cdot 10^{-6}$ &         234.887 &         0.815 &            0.816 &        0.111 &            0.026 \\
    \midrule
    \multirow{2}{*}{\textbf{Minnesota}} & \textbf{Density} &  9.47$\cdot 10^{-4}$ &             0.999 &                0.999 &                0.010 &           0.999 &         0.002 &            0.002 &        0.003 &            0.002 \\
            & \textbf{Median} &                1.000 &             0.004 &  7.58$\cdot 10^{-4}$ &                0.014 &           0.013 &         1.000 &            1.000 &        0.333 &            0.204 \\
    \midrule
    \multirow{2}{*}{\textbf{Sensor}} & \textbf{Density} &                0.159 &             0.969 &                0.969 &                0.844 &           0.875 &         0.273 &            0.230 &        0.529 &            0.227 \\
            & \textbf{Median} &                0.742 &             0.463 &  2.42$\cdot 10^{-4}$ &                0.005 &           6.147 &         0.765 &            0.756 &        0.201 &            0.103 \\
    \midrule
    \multirow{2}{*}{\textbf{Airfoil}} & \textbf{Density} &                0.001 &             1.000 &                1.000 &                0.009 &           0.996 &         0.003 &            0.003 &        0.006 &            0.002 \\
            & \textbf{Median} &                0.500 &             0.003 &  4.70$\cdot 10^{-4}$ &                0.026 &           0.209 &         0.500 &            0.500 &        0.090 &            0.118 \\
    \bottomrule
    \end{tabular}
    }
\end{table}

\end{document}